\documentclass[runningheads]{llncs}
\usepackage[T1]{fontenc}

\usepackage{times}

\usepackage{cite}
\usepackage{multicol}
\usepackage[bookmarks=true, colorlinks=true, urlcolor=orange]{hyperref}
\usepackage[utf8]{inputenc}

\usepackage{booktabs}
\usepackage{amssymb}

\usepackage{makecell}

\usepackage{pifont}
\newcommand{\cmark}{\ding{51}}  %

\usepackage{xcolor}
\definecolor{rrt}{HTML}{A01CBB}
\definecolor{faded1 rrt}{HTML}{EE99CC}

\definecolor{eit}{HTML}{009E1A}
\definecolor{faded1 eit}{HTML}{90E93D}

\definecolor{ait}{HTML}{00C3FF}
\definecolor{faded1 ait}{HTML}{306DDF}

\definecolor{prioritized}{HTML}{FFD61F}

\definecolor{prm}{HTML}{E21616}
\definecolor{faded1 prm}{HTML}{FF6600}

\definecolor{birrt}{rgb}{0, 0, 0}
\definecolor{faded1 birrt}{HTML}{9e9aa1}

\definecolor{rnd_sample}{HTML}{1f77b4}
\definecolor{edge_valid}{HTML}{2ca02c}
\definecolor{edge_invalid}{HTML}{d62728}
\definecolor{conf_check}{HTML}{ff7f0e}

\usepackage{etoolbox}
\usepackage[linesnumbered,vlined,ruled,commentsnumbered]{algorithm2e}
\SetKwRepeat{Do}{do}{while}

\SetVlineSkip{0.2em}

\SetInd{0.2em}{0.7em}

\DontPrintSemicolon{}

\SetKwIF{If}{ElseIf}{Else}{if}{}{else if}{else}{endif}

\SetAlgoSkip{}

\SetAlCapHSkip{0cm}

\SetAlgoCaptionLayout{small}
\SetAlFnt{\small}

\SetArgSty{textnormal}

\SetKwComment{Comment}{$\triangleright$\ }{}

\makeatletter
\patchcmd\algocf@Vline{\vrule}{\vrule \kern-0.4pt}{}{}
\patchcmd\algocf@Vsline{\vrule}{\vrule \kern-0.4pt}{}{}
\makeatother

\usepackage{tikz}

\usepackage{tikzscale}

\usepackage{pgfplots}

\usepackage{pgfplotstable}

\usetikzlibrary{calc}

\usetikzlibrary{positioning}

\usepgfplotslibrary{statistics}

\usepgfplotslibrary{fillbetween}

\pgfplotsset{
  compat=1.15,
  mps basic/.style={
    xlabel near ticks,
    xlabel style={font=\footnotesize},
    ylabel near ticks,
    ylabel style={font=\tiny},
    xmajorgrids,
    major x grid style={dotted},
    ymajorgrids,
    major y grid style={dotted},
    tick label style={font=\tiny}
  },
  mps scientific x/.style={
    x tick label style={
      /pgf/number format/sci,
      font=\tiny
    }
  },
  mps scientific y/.style={
    y tick label style={
      /pgf/number format/sci,
      font=\tiny
    }
  },
  mps fixed x/.style={
    x tick label style={
      /pgf/number format/.cd,
      fixed,
      fixed zerofill,
      precision=6,
      /tikz/.cd,
      font=\tiny
    }
  },
  mps fixed y/.style={
    y tick label style={
      /pgf/number format/.cd,
      fixed,
      fixed zerofill,
      precision=6,
      /tikz/.cd,
      font=\tiny
    }
  },
  /pgfplots/myylabel absolute/.style={%
      /pgfplots/every axis y label/.style={at={(0,0.5)},xshift=#1,rotate=90,align=center},
      /pgfplots/every y tick scale label/.style={
        at={(0,1)},above right,inner sep=0pt,yshift=0.3em
      }
   }
}

\usepackage{xcolor}
\definecolor{espblack}{RGB}{0,0,0}
\definecolor{espwhite}{RGB}{255,255,255}
\definecolor{espgray}{RGB}{206,206,206}
\definecolor{esplightgray}{RGB}{224,224,224}
\definecolor{espdarkgray}{RGB}{168,168,168}
\definecolor{espsomewhatdarkgray}{RGB}{130,130,130}
\definecolor{espverydarkgray}{RGB}{100,100,100}
\definecolor{espblue}{RGB}{11,93,174}
\definecolor{esplightblue}{RGB}{59,175,236}
\definecolor{espdarkblue}{RGB}{6,26,64}
\definecolor{espred}{RGB}{206,62,21}
\definecolor{esplightred}{RGB}{206,62,21}
\definecolor{espdarkred}{RGB}{61,19,8}
\definecolor{espyellow}{RGB}{232,163,26}
\definecolor{espgreen}{RGB}{100,161,27}
\definecolor{esplightgreen}{RGB}{149,198,35}
\definecolor{espdarkgreen}{RGB}{49,92,43}
\definecolor{esppurple}{RGB}{106,20,125}
\definecolor{esplightpurple}{RGB}{197,137,232}
\definecolor{espdarkpurple}{RGB}{50,14,59}

\usepackage{amsmath}
\usepackage{mathrsfs}
\usepackage{mathtools}
\usepackage{amsfonts}

\usepackage{subcaption}
\usepackage{pgfplots}

\usepackage{dblfloatfix}

\usepackage[capitalise]{cleveref}

\usepackage{siunitx}

\usepackage{appendix}

\usepackage{multirow}

\usepackage{arydshln}

\usepackage{stackengine}

\usepackage[export]{adjustbox}

\title{ \bf
Sampling-Based Multi-Modal Multi-Robot Multi-Goal Path Planning%
}

\makeatletter
\newcommand{\printfnsymbol}[1]{%
  \textsuperscript{\@fnsymbol{#1}}%
}
\makeatother

\author{Valentin N. Hartmann\thanks{equal contribution}, Tirza Heinle\printfnsymbol{1}, Yijiang Huang, Stelian Coros}

\authorrunning{V. N. Hartmann et al.}

\institute{Computational Robotics Lab\\
ETH Zürich, Zürich, CH}

\newcommand{\arxiv}[1]{#1}

\begin{document}
\maketitle

\begin{abstract}

In many robotics applications, multiple robots are working in a shared workspace to complete a set of tasks as fast as possible.
Such settings can be treated as multi-modal multi-robot multi-goal path planning problems, where each robot has to reach \textit{a set} of goals.
Existing approaches to this type of problem often use some type of decomposition, and many are neither optimal nor complete.
Given the lack of optimal solvers for this problem, it is unknown how big the optimality gap of current solvers is.

We formalize this problem as a centralized path planning problem and present planners that are probabilistically complete and asymptotically optimal.
The planners are modifications of standard sampling-based planners with the required changes to work efficiently in the multi-modal, multi-robot, multi-goal setting.
We validate the planners on a diverse range of problems including scenarios with various robots, planning horizons, and collaborative tasks such as handovers, and compare the planners against a suboptimal prioritized planner.
With the optimal planners, we answer the question of how close to optimal the the suboptimal prioritized planners are.
Videos and code for the planners and the benchmark is available at \url{https://vhartmann.com/mrmg-planning/}.

\end{abstract}

\section{Introduction}

As adoption of robots increases and tasks become more and more automated, it will be increasingly important to deploy solutions that are not only able to work longer and cheaper but are also competitive in throughput with humans.
In order to achieve this, in many cases, multiple robots need to be used and effectively coordinated in the same workspace: Enabling motion planning for multiple tasks with multiple robots is crucial in order to maximize the usefulness of robots in industrial settings.
While workcells with multiple robots exist, the robots typically act independently from each other in order to simplify the programming and avoid dealing with robot-robot interactions.

Most work in continuous multi-robot planning is focusing on single-goal settings where all robots start moving at the same time and reach their respective goals simultaneously \cite{wagner2015subdimensional,shome2021asymptotically,lin2022review}.
Conversely, in most real use cases, robots need to do multiple tasks in sequence, e.g., welding multiple points, or picking and placing multiple things after another in order to sort objects.
Even when only considering a single pick and place task per robot, each robot needs to reach two goals: The pick, and the place location.
Since the robots act in the same environment in these scenarios and thus possibly block each other from doing their tasks, we can not formulate the problem at hand as a \textit{sequence of path planning problems}, but need to solve the multi-robot multi-goal planning problem if we want to find an optimal solution.

Multi-goal planning can also be framed as multi-modal planning:
Multi-modal path planning \cite{thomason_ral22, hauser2010task, englert2021rss,schmitt2017optimal} finds paths through sequences of \textit{modes}, i.e., through variations of a continuous configuration space that occur, e.g., by grasping an object, or moving an object in the workspace.
This means that in each \textit{mode} a different set of constraints is active.
Most work on multi-modal path planning for robots only considers single-robot settings.

\begin{figure}[t]
    \centering
    \includegraphics[width=0.3\linewidth]{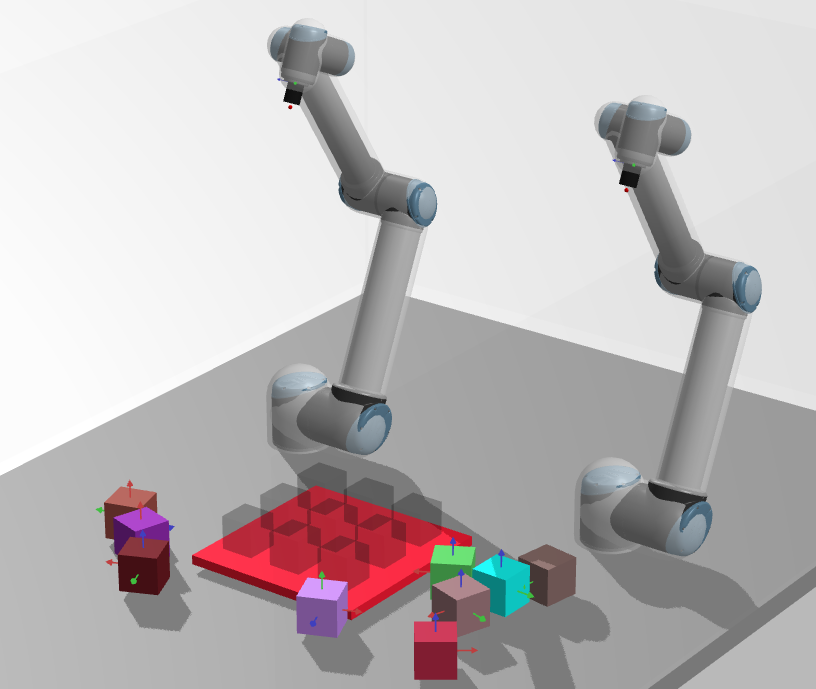}
    \includegraphics[width=0.3\linewidth]{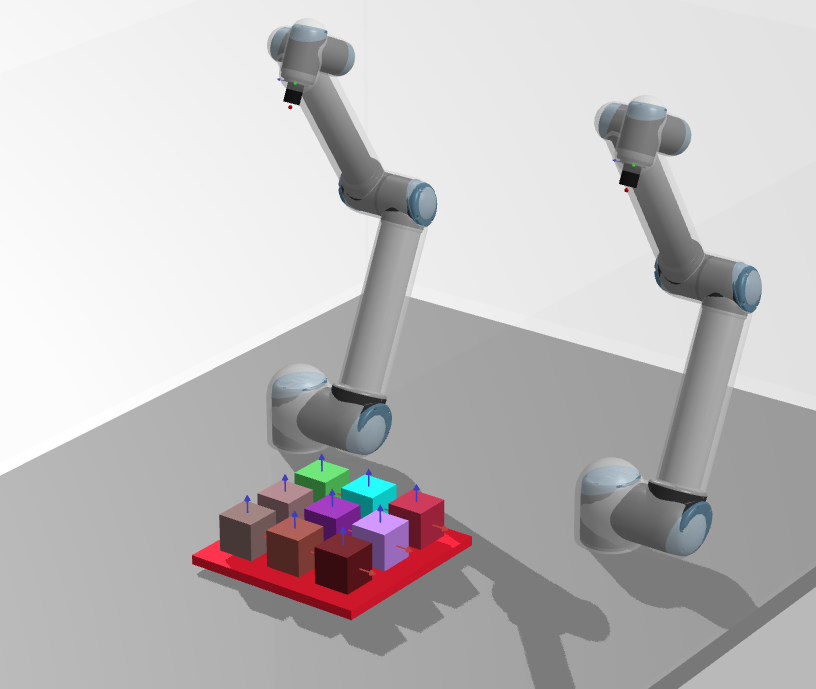}\\
    \vspace{1mm}
    \includegraphics[width=0.3\linewidth]{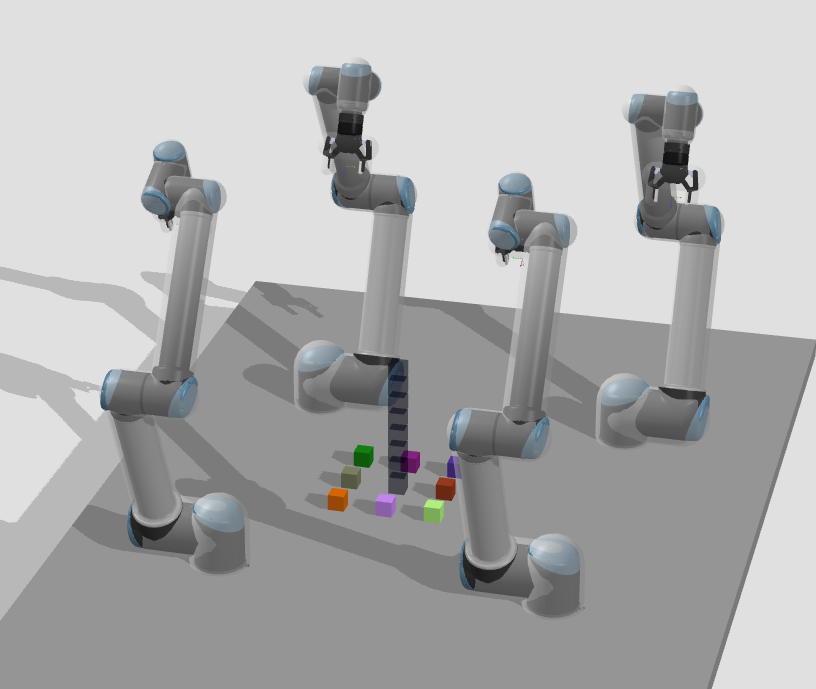}
    \includegraphics[width=0.3\linewidth]{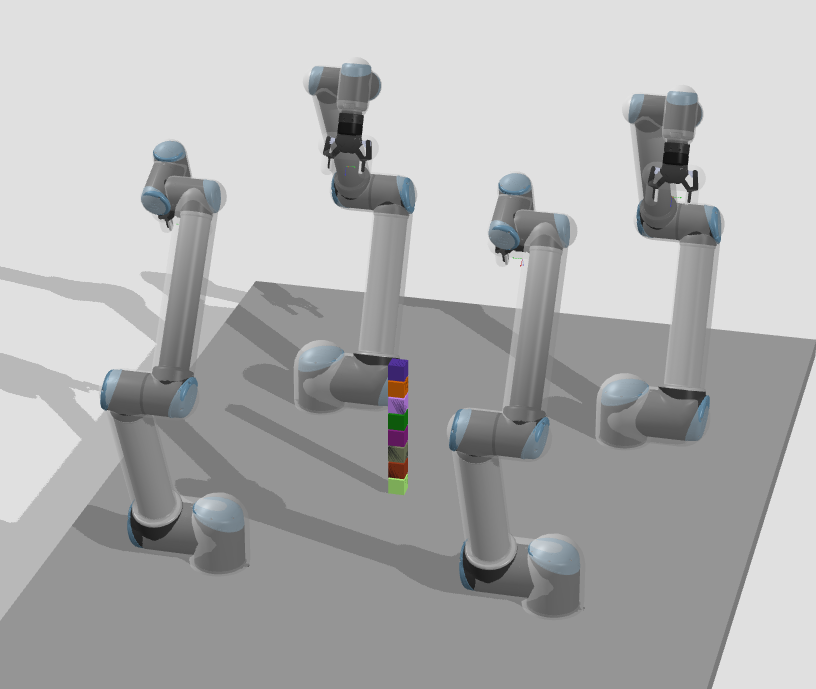}
    \caption{Examples of problems with their initial state (left) and goal state (right): (top) two robots cooperating to reorient boxes, (bottom) four robots stack boxes on top of each other.}
    \label{fig:intro_image}
\end{figure}

For the multi-robot, multi-goal planning problem we consider, only suboptimal planners exist \cite{hartmann2022long, 23-hartmann-robplan}.
In this paper, we propose complete and optimal planners for the multi-agent, multi-goal, multi-modal planning problem, and with the help of these planners both provide optimal baselines for the problem, and answer the question \textit{'How close to optimal are the suboptimal planners?'}. 
Summarizing, our contributions in this paper are

\begin{itemize}
    \item a formalization of multi-modal, multi-robot, multi-goal path planning in continuous spaces,
    \item adaptations of probabilistically complete and almost-surely asymptotically optimal planners to this setting.
    
\end{itemize}
We compare the planners to a suboptimal, incomplete, prioritized planner on a multi-modal, multi-robot, multi-goal motion planning benchmark containing various base-scenarios that can be further adjusted in difficulty by changing the number of robots and tasks.

\section{Related Work}

\begin{table}[t]
    \centering
    \caption{Comparison of path planning approaches. Ticks in parentheses mean that the approach is only optimal/complete under some assumptions that do not always hold.}
    \scriptsize
    \setlength{\tabcolsep}{3pt}
    \renewcommand{\arraystretch}{1.0}
    \begin{tabular}{l@{\hspace{4pt}}lcclccc}
        \toprule
        Planning Space & Approach & Examples & \makecell{Multi-\\Robot} & \makecell{Task\\Structure} & \makecell{Multi-\\Modal} & \makecell{Complete} & Opt. \\
        \midrule

        \multirow[t]{2}{*}{Discrete} 
        & MAPF & \cite{stern2019multi,yu2016optimal} & \cmark & Single goal &  & \cmark & \cmark \\
        \cmidrule(l){2-8}
        & MAPD & \cite{ma2017lifelong,grenouilleau2019multi} & \cmark & Goal sequence &  & \cmark & \cmark \\
        \hline

        \multirow[t]{4}{*}{Continuous} 
        & Multi-Modal Planning & \cite{thomason_ral22, hauser2010task} &  & General task graphs & \cmark & \cmark & \cmark \\
        \cmidrule(l){2-8}
        & MRMP & \cite{moldagalieva2024db,shome2020drrt,02-sanchez,orthey2024multilevel} & \cmark & Single goal & \cmark & \cmark &  \\
        \cmidrule(l){2-8}
        & Prioritized Planning & \cite{hartmann2022long,chen2022cooperative} & \cmark & Goal sequence & \cmark & (\cmark) &  \\
        \cmidrule(l){2-8}
        & Hybrid Methods & \cite{motes2023hypergraph} & \cmark & Unordered goals & \cmark & \cmark & (\cmark) \\
        \cmidrule(l){2-8}
        & Ours &  & \cmark & General task graphs & \cmark & \cmark & \cmark \\
        \bottomrule
    \end{tabular}
    \label{tab:comparison}
\end{table}

Most multi-robot planning work is focusing either on the setting where all robots share the same work and configuration space, or on solving single goal problems.
It is commonly assumed that the environment stays completely static during planning, therefore excluding manipulation planning settings.
We group the different research areas and their respective focus in \cref{tab:comparison}.
In the following, we discuss single-goal multi-robot planning (MRMP) and multi-agent path finding (MAPF) more in-depth.

\subsection{Continuous Multi-Robot Planning}

A simple approach to multi-robot planning is planning in the composite space of all robots \cite{02-sanchez}, and applying standard methods such as RRT(*) \cite{lavalle2001rapidly} or PRM(*) \cite{kavraki1996probabilistic} in this higher dimensional space.
This approach is not scalable to many robots or high degrees of freedom.

There are various approaches to tackle this limitation.
Discrete RRT (dRRT(*) \cite{solovey2016finding,shome2020drrt}), builds a roadmap for each robot, and combines them to implicitly form a tensor graph.
Instead of planning naively on the implicit graph, an RRT-like strategy is proposed to explore this implicit graph.
Fast-dRRT \cite{solano2023fast} applies this approach sequentially on the subproblems to solve multi-goal problems.
M* \cite{wagner2015subdimensional} plans in separate spaces, and plans jointly only where required, i.e., when conflicts between the single-robot plans arise.
Similarly, conflict based search (CBS) \cite{sharon2015conflict} plans for each robot individually, and introduces constraints to solve conflicts between single agent plans.
CBS originates from the MAPF-setting, but can also be applied to continuous multi-robot planning problems \cite{solis2021representation, kottinger2022conflict, moldagalieva2024db}.

Prioritized planners as proposed in, e.g., \cite{orthey2024multilevel, hartmann2022long, van2005prioritized,chen2022cooperative} achieve more scalable solvers by not requiring completeness and optimality.
Similarly, \cite{okumura2022quick} proposes a scalable approach that is based on operator splitting and sampling, and while the approach is complete, it is not optimal.

The prioritized planners can be used to plan for multi-goal problems by incrementally generating a path per agent, and considering it as fixed for the later planning problems as done in \cite{chen2022cooperative,hartmann2022long}.
To the best of the authors' knowledge, there are no multi-modal multi-robot multi-goal planners that are complete and optimal in the general setting.

A slightly different approach is taken in \cite{han2024stop}, where plans for the separate robots are planned, and then - if conflicts between the paths arise - fixed by introducing pauses.

Compared to the discussed works, we are interested in optimally solving multi-robot multi-goal planning problems without any restricting assumptions.
Further, we are interested in manipulation, i.e., problems where the environment changes through the actions of the robot.

\subsection{Multi-Agent Path Finding}
MAPF \cite{stern2019multi} generally refers to planning in grid-like 2D environments with disk-robots, as found in warehouses.
A common approach to solve this problem is conflict based \cite{sharon2015conflict} or priority based search (PBS) \cite{ma2019searching}.
Many works improve upon the initial approach, such as improved CBS \cite{boyarski2015icbs}, or bounded suboptimal CBS \cite{li2021eecbs}, scaling to 1000s of agents.

Multi-agent pickup and delivery (MAPD) refers to the setting where we do not only intend to go from point A to point B, but traveling between multiple points of interest, as part of a single planning problem \cite{ma2017lifelong, salzman2020research}.
While MAPD can be framed as lifelong MAPF problem, i.e., once an agent has finished its task of the MAPF problem, a new goal is assigned to the robot, these solvers are suboptimal \cite{mouratidis2024fools}.
Multi-label A* \cite{grenouilleau2019multi} extends the classic A* algorithm to a sequence of multiple goals, and is able to optimally solve MAPD problems.
Both MAPF and MAPD planners typically assume that the environment remains unchanged over the planning duration.

\section{Multi-Robot Multi-Goal Path Planning}
Informally, our objective is to find a collision free path for each robot that passes through a sequence of goals that minimizes a cost (e.g., the latest completion time of all robots).
Goals might involve multiple robots: A handover of an object between two robots implies a constraint on two robots.
A goal could additionally imply a \textit{mode transition}, i.e., a robot grasping something and thus changing the environment for the remaining planning problem.

\subsection{Preliminaries}
\begin{figure}[t]
    \centering
    \includegraphics[width=0.8\linewidth]{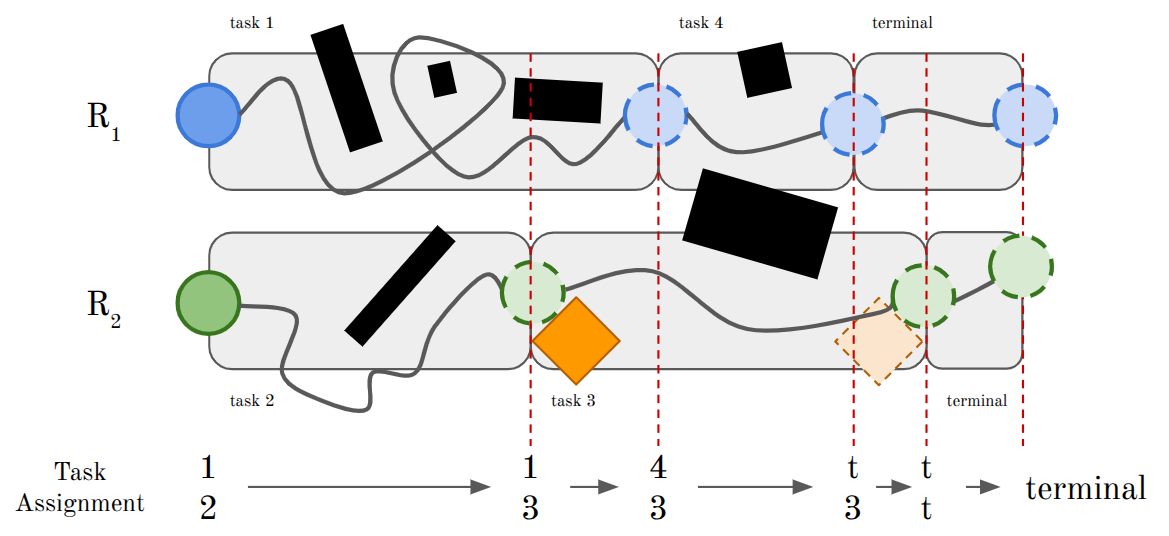}
    \caption{An illustration for what tasks are, and how robots move through task assignments and space together. The robots start with the task assignment $[1, 2]^T$ and proceed to the terminal mode. The goals of the tasks are shown with lower opacity and dashed lines. The vertical red dashed lines indicate points where a task is completed, and the task assignment changes. Task $t$ indicates the terminal task, that both robots need to reach jointly.}
    \label{fig:illustration}
\end{figure}
To formalize the problem, we first introduce the notion of tasks, the concept of a mode, and the space that we plan in.
We follow the work from Thomason et al. \cite{thomason_ral22}, and generalize it to multiple robots.
Compared to \cite{thomason_ral22}, task sequencing and allocation happens implicitly in the planning.
Refer to \cref{fig:illustration} for an informal schematic of what tasks are, and how the task assignment changes over time.

\subsubsection{Task}
A task consists of the robots that are assigned to the task, and a set of goal-constraints $g$ that need to be fulfilled to consider the task done\footnote{Additionally, a task could include path constraints $h$ during a task, such as, e.g., an orientation constraint on the end-effector of a robot. This requires constrained planning, which we leave for future work.}.
Typically, the constraint $g$ is a single goal pose or a goal region, but $g$ could also be a more complex nonlinear constraint such as a grasp constraint.
We do not assume anything on how $g$ is defined except that we can sample constraint-satisfying configurations (e.g., using projection methods).
A task can have post-conditions that can alter the scene-graph of the environment, i.e., which objects are linked to each other. 
As example, consider the task `robot $r_1$ grasps object $o_1$': Here, the goal constraints are that the robot is grasping $o_1$, and the post-condition is that $o_1$ is linked to the end-effector of the robot.

We use $s \in \mathcal{S} = \mathcal{S}_{r_1}\times...\times \mathcal{S}_{r_n}$ to denote the \textit{task assignment} of all robots, where $\mathcal{S}_{r_i}$ is the set of tasks that can be assigned to robot $r_i$.

\subsubsection{Modes}
The constraints of a task can be fulfilled by a set of poses, where the chosen pose might affect the environment that we plan in due to a tasks post-condition.
Consider again the grasping-task: The robot is able to fulfill a grasping constraint by grasping from any side, which changes the collision geometry that we need to consider in the rest of the planning problem. %
We use \textit{mode} $m\in\mathcal{M}=\mathcal{S}\times\mathcal{Q}_o$
to refer to the combination of the discrete task assignment $s$ (which implies a scene-graph, i.e., which objects are linked to which parents), and the poses of all movable objects $q_o\in\mathcal{Q}_o$, defined by the relative transformation to their parent-frames.
Importantly, defining $q_o$ as relative transformation to the parent frame means that a mode \textit{only changes} when the scene-graph changes, i.e., upon completion of a task that has post-conditions that change the scene-graph.

\begin{figure*}[t]
    \centering
    \includegraphics[width=1\linewidth]{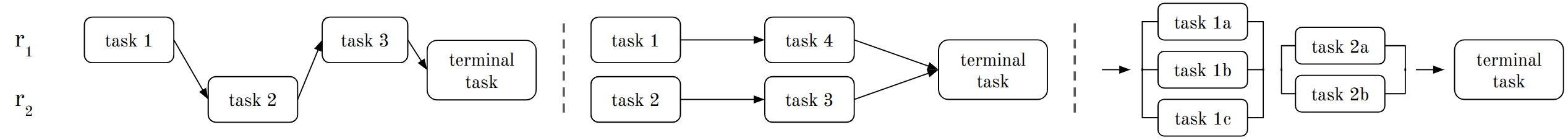}
    \caption{Examples for the different ways in which a task ordering can be specified: \textbf{(left)} As a sequence of tasks \textbf{(middle)} as a partially ordered sequence and \textbf{(right)} as a set of unassigned and unordered tasks (i.e., the order of task groups 1 and 2 is exchangeable).}
    \label{fig:sequence_dependency_graph}
\end{figure*}

\begin{figure}[t]
    \centering
    \includegraphics[width=0.7\linewidth]{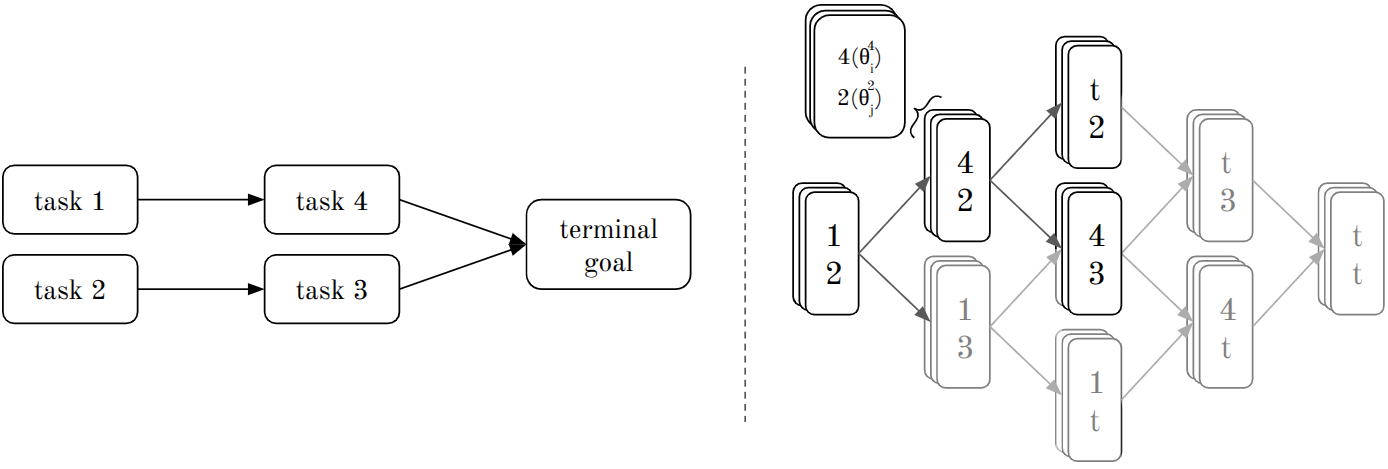}
    \caption{Example of a mode-graph ($\mathcal{G}(\mathcal{O})$, right) that is implied by the ordering on the left. Modes with full opacity show an example of how a set of modes can be incrementally reached. Each mode can be reached with different parameterizations $\theta_i^{\text{k}}$ (e.g., different grasps), as indicated for mode (4, 2). \cref{fig:illustration} is one possible realization for this problem specification.}
    \label{fig:mode_graph}
\end{figure}

\subsubsection{Task Order Specification}
A task order is induced by an oracle $\mathcal{O}(m)\subseteq\mathcal{S}$ which, given the current mode, returns a set of \textit{all} possible successor task assignments.
Note that the task assignment (and thus the mode) only changes when the goal constraints of a previous task are fulfilled.
Some examples of how a task order can be specified are illustrated in \cref{fig:sequence_dependency_graph}.
The oracle induces a graph $\mathcal{G}(\mathcal{O})$ of task assignments that we can traverse to go from the start to the end-mode (\cref{fig:mode_graph}, right).

\subsubsection{Configuration Space}
We describe a path in the composite space of all robots, all objects, and which tasks are currently assigned to each robot:
\begin{equation} \label{eq:conf_space}
    \mathcal{Q} = \underbrace{\mathcal{Q}_{r_1} \times\cdots\times \mathcal{Q}_{r_N}}_{\mathcal{Q}_R} \times \underbrace{\mathcal{S}_{r_1} \times\cdots\times \mathcal{S}_{r_N}}_{\mathcal{S}} \times \mathcal{Q}_o.
\end{equation}
Here, $\mathcal{Q}_{r_i}$ is the configuration space of robot $r_i$, and correspondingly, $\mathcal{Q}_R$ is the composite configuration space of all robots; $\mathcal{Q}_o$ is the composite configuration space of all objects. %
We use $\mathcal{Q}_\text{free}\subseteq\mathcal{Q}$ to denote the part of the configuration space that is collision-free.

Clearly, not all degrees of freedom are actuated.
We assume that we can only plan for the robots' degrees of freedom directly, and all others need to be influenced indirectly.
Particularly, the object configurations and the task assignment can only be changed through the completion of tasks (i.e., fulfilling the goal constraints $g$ of a task).

\subsection{Problem formulation}
A multi-modal, multi-robot, multi-goal path planning problem is then given by the tuple $(R, \mathcal{Q}_R, q_\text{start}, m_\text{start}, \mathcal{O})$, where $R$ is the set of robots, $\mathcal{Q}_R$ is the configuration space of the robots, $q_\text{start}$ is the initial configuration, $m_\text{start}$ is the start mode, and $\mathcal{O}$ is the oracle giving us the possible next task assignments from the current mode.
In the following, we will use $q^{r_i}$ for the pose of robot $r_i$.

We want to find a collision free path $\pi(t): \mathbb{R} \to \mathcal{Q}_R$ and the task assignment sequence $s(t): \mathbb{R} \to \mathcal{S}$, that minimizes the cost function $c(\cdot)$.
The path $\pi$ maps time to the composite robot configuration $\mathcal{Q}_R$, and the mode sequence $s(t)$ maps time to the task assignment $\mathcal{S}$, which together imply the scene-graph at a time, and thus the poses of all objects.

\subsubsection*{Cost functions}
We are often interested in finding the minimum makespan plan.
However, purely minimizing makespan can lead to optimal paths that bring undesired side-effects, e.g., containing unnecessary movement, due to the large nullspace in the makespan cost.
Thus, we consider cost functions of the family
\begin{equation} \label{eq:cost_fct}
c(q_1, q_2) = (1\!-\!w) \max_r ||q^r_1\!-\!q^r_2||_2 + w \sum\nolimits_{r} ||q^r_1\!-\!q^r_2||_2.
\end{equation}
where if $w$ is small, we get a minimum makespan optimization problem (which is `regularized' by the path length), and if $w$ is $1$ the total cost is the sum of all robot path lengths.

\section{Planners}

We now introduce the necessary adaptations to four sampling-based planners (RRT*, PRM* \cite{karaman2011sampling}, AIT* and EIT* \cite{strub_ijrr22}).
We refer to the original works for a detailed explanation of how the algorithms work.
Generally, the planners proceed as their standard version, with an adjusted distance function that accounts for the connectivity between the modes.
Before discussing any specific adjustments required in the planners in \cref{sec:considerations}, we discuss mode sampling, the distance function used in the planners, and the concept of the informed set \cite{gammell_iros14}.

\subsubsection{Mode sampling}
Instead of sampling the (unactuated) object poses and task assignments, we sample $x=(q, m)$, where $q\in\mathcal{Q}_R$ is a configuration containing all robots' configurations and $m$ is a mode that determines the task assignment and all object poses.
To sample the modes, we do not sample the continuous space $\mathcal{S}\times\mathcal{Q}_O$, but we sample from the list of previously reached modes.
To enable that, we store the list of modes $M$ that we reached so far, and expand it whenever we reach a new mode (i.e., whenever we reach a configuration that fulfills the constraints $g$ of a task).
To sample $m$ from the set $M$, we maintain the most recently added mode $F_{-1}$ and the set of modes $F$ that we did not yet expand (the frontier).
We then sample according to:
\begin{equation*}
\label{eq:prob_dist}
m \sim 
\begin{cases}
\mathcal{U}(F) & \text{with probability } p(1 - \epsilon) \\
\mathcal{U}(F_{-1}) & \text{with probability } p\epsilon \\
\mathcal{U}(M \setminus F) & \text{with probability } 1 - p
\end{cases}
\end{equation*}
where $p \in [0,1]$ and $\epsilon \in [0,1]$ are weighting parameters, and $\mathcal{U(\cdot)}$ denotes a uniform distribution over a finite set.
If $p = 1$ and $\epsilon = 1$, we reach a fully \textit{greedy} version, resulting in the most direct progression possible towards the terminal mode. 

We refer to the \textit{frontier sampling strategy} if $p>0.5$ holds. 
This biases the planner towards the unexpanded modes (i.e., frontiers) compared to the fully greedy approach, while still maintaining a greedy progression toward the terminal state\footnote{A fully greedy strategy is not probabilistically complete, as the planners might get stuck in a dead-end.}.

\subsubsection{Distance function}
We assume that any node in a different mode is inaccessible except through the transition nodes:
$$
d(x_1, x_2) = 
\begin{cases} 
d_{\text{mode}}(q_1, q_2) & \text{if } m_1 = m_2, \\
0 & \text{if } q_1 = q_2 \text{ and } s(m_2)\in\mathcal{O}(m_1) \text{ and } g(q_1, m_1) = 0\\
\infty & \text{otherwise},
\end{cases}
$$
where we use $s(m)$ to denote the task assignment of mode $m$.
For the in-mode distance any metric can be chosen; the work in  \cite{atias2018effective} discusses distance metrics in multi-robot problems in more depth.
We use the maximum of the per-robot $L_2$ distance, as this performed best in our experiments.

\subsubsection{Locally informed sampling}
Once we found a solution, the informed set \cite{gammell_iros14} can be used to restrict the points that we need to consider for planning to the points that can improve the current solution.
The informed set in our problem is 
\begin{equation}
    \mathcal{Q}_\text{inf} = \{q\, |\, c_\text{lb}(q_\text{start}, q) + c_\text{lb}(q, Q_\text{goal}) \leq c_\text{best}\},\label{eq:infset}
\end{equation}
with $Q_\text{goal}$ denoting the set of possible goal configurations.

In our setting, the optimal paths are long compared to the size of the configuration space.
Therefore, the informed set can be larger than the valid configuration space%
, and thus not helpful.

Given that we do know that a solution needs to pass through some regions in space (the goals that we need to reach), the standard approach of taking the cost \cref{eq:cost_fct} as $c_\text{lb}$ is not approximating the true path cost well across modes.
It is not easy to construct a good lower bound for the path-cost, since the regions that we need to pass through are \textit{subspaces of the full space}: even though we know that robot $r_i$ needs to be at pose $q_{g_1}$ first, robot $r_j$ might be unconstrained, thus resulting in a volume that we need to pass through instead of a single point.
Additionally, compared to the Euclidean norm as cost, our cost function does not easily allow for direct sampling from the informed set. %

To alleviate these issues, we propose \textit{locally informed sampling}: instead of using the start and the goal in \cref{eq:infset}, we randomly sample two points $q_i, q_j$ on the previously found path, and do informed sampling with these points:
\begin{equation*}
    \mathcal{Q}_\text{lis} = \{q\, |\, c(q_i, q) + c(q, q_j) \leq c^\pi(q_i, q_j)\},\label{eq:locally_infset}
\end{equation*}
where $c^\pi(\cdot)$ is the cost to reach a point on the path along the best found path. %
We extended this to the sampling of modes by sampling from the set of modes that are in-between the modes that $q_i, q_j$ are in respectively, i.e., the modes in $\mathcal{G}(\mathcal{O})$ that are reachable from $m_i$ and can reach $m_j$.
Locally informed sampling does not influence optimality or completeness guarantees, since we sample the indices randomly, and thus also sample from the standard informed set.

\subsection{Considerations for Multi-Modal, Multi-Goal Planners}\label{sec:considerations}

\subsubsection{RRT*}\label{sec:rrt}
Compared to standard RRT* \cite{lavalle2001rapidly,karaman2011sampling}, we maintain a set of previously reached modes $M$ (i.e., a mode-graph similar to \cref{fig:mode_graph}), which is initialized with the start mode.
In order to sample configurations in the planner, we uniformly sample poses within the limits of the robots, and sample modes from this set.
If we previously found a path, we use locally informed sampling to sample the mode and the pose.
When using a goal bias, we bias the sampling in the RRT to the \textit{goal in a mode} (i.e., to a configuration satisfying the goal constraints $g$ of a task from the current task assignment).

We also use a bidirectional variant (BiRRT*), which first samples goals in each mode, which are used as starts in the next mode, and then runs bidirectional RRT* in each mode.

\subsubsection{PRM*}\label{sec:prm}
The main change in the PRM* planner is that the roadmap is constructed by first sampling a batch of transition nodes (i.e., nodes that fulfill the constraints of a task), which enlarges the set of reached modes $M$ and then sampling a batch of nodes in the reached modes, until the terminal mode is reached.
Then, a standard A* with an edge queue is used for the search over the multi-modal roadmap.

\paragraph*{Heuristics}
Since A* needs a heuristic in order to be effective, and the Euclidean distance is not informative in our setting, we compute a heuristic for the A* search by running a reverse search on the transition nodes to compute a lower bound on the cost to reach a mode.
The heuristic is then the sum of the cost to reach a specific transition and the cost to reach the goal from this transition mode:
\begin{equation}\label{eq:heuristic}
    h(x=(q, m)) = \min_i\left[c(q, q_i^{m,\text{trans}}) + c_\text{rev}(q_i^{m, \text{trans}})\right]
\end{equation}
where $q^{m,\text{trans}}$ are the poses belonging to the transition nodes in a mode.
This heuristic is always underestimating the true cost, and is thus admissible, and since $c$ and $c_\text{rev}$ are consistent, $h$ is also consistent.

\subsubsection{Informed Tree-planners}\label{sec:it}
AIT* and EIT* \cite{strub2020adaptively, strub_ijrr22} use adaptive problem-specific heuristics to focus the search on promising regions.
These heuristics are computed and refined through an \textit{asymmetric bidirectional search} to obtain more informative exploration and improved solution quality.
\paragraph{Search} To explore the graph, AIT* and EIT* perform their respective \textit{asymmetric bidirectional search} strategies, as described in \cite{strub_ijrr22}. 
Although each planner utilizes distinct heuristics, both follow an identical approach to handle transitions: When the forward search reaches a transition node, it only expands to valid successor modes, enforcing forward-directed connectivity; the reverse search applies the same logic in opposite direction (i.e., expanding towards previous modes). 
Both planners need an admissible a priori cost-to-come (i.e., from the start to the state $x$) heuristic - we use a modified version of \cref{eq:heuristic} with the lower bound cost to reach a mode computed from the start node instead of the goal (i.e., replacing $c_\text{rev}$ with the cost computed by running a forward search only over the transition nodes of the previous modes).
This heuristic is a trade-off between heuristic accuracy and computational efficiency.

The reverse search in EIT* incorporates heuristics approximating the computational effort required to find a valid path. 
We define the computational effort as the collision checking effort.
Following this definition, the effort-to-come is computed analogously to the cost-to-come heuristic in \cref{eq:heuristic}, where cost terms are substituted by effort.

\begin{figure*}[t]
    \begin{subfigure}[t]{.16\textwidth}
        \centering
        \includegraphics[width=.97\linewidth]{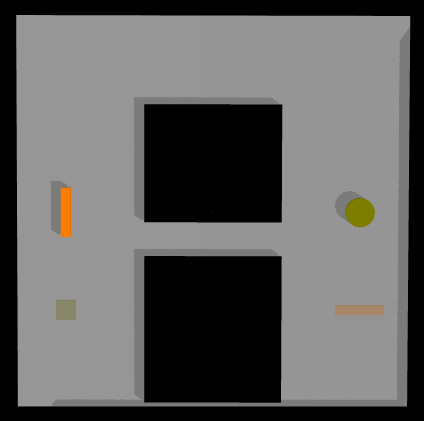}
        \caption{\label{fig:hallway}2D hallway.}
    \end{subfigure}\hfill
    \begin{subfigure}[t]{.16\textwidth}
        \centering
        \includegraphics[width=.97\linewidth]{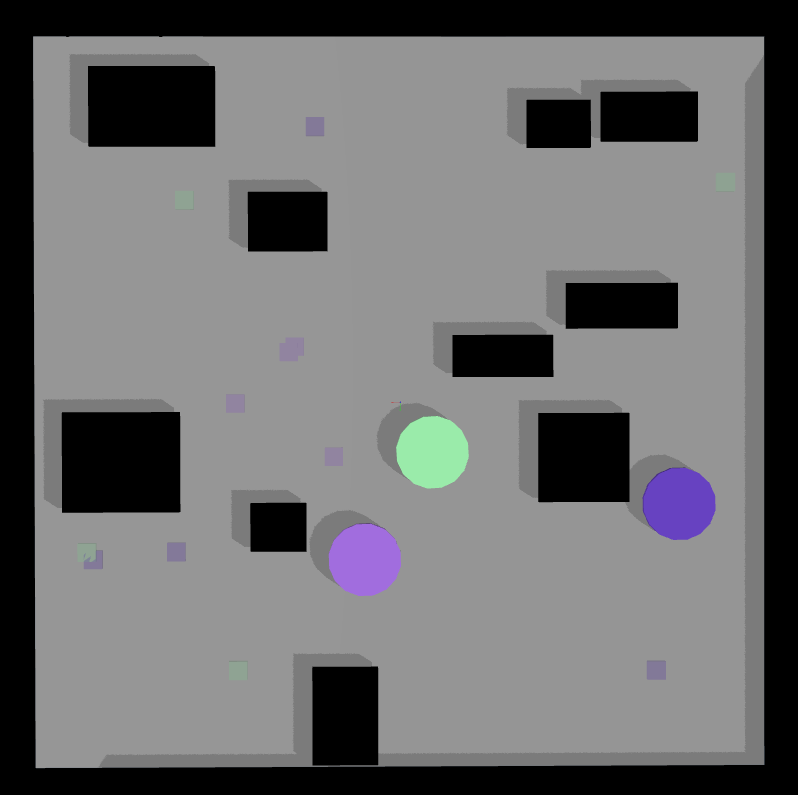}
        \caption{\label{fig:random} 2D random.}
    \end{subfigure}\hfill
    \begin{subfigure}[t]{.16\textwidth}
        \centering
        \includegraphics[width=.97\linewidth]{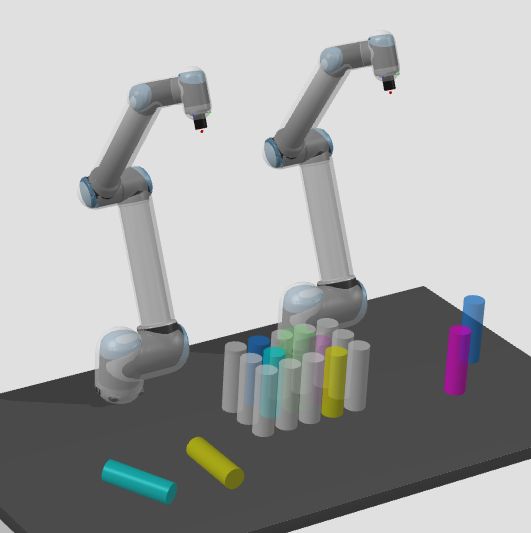}
        \caption{Bottle insertion.}
    \end{subfigure}\hfill
    \begin{subfigure}[t]{.16\textwidth}
        \centering
        \includegraphics[width=.97\linewidth]{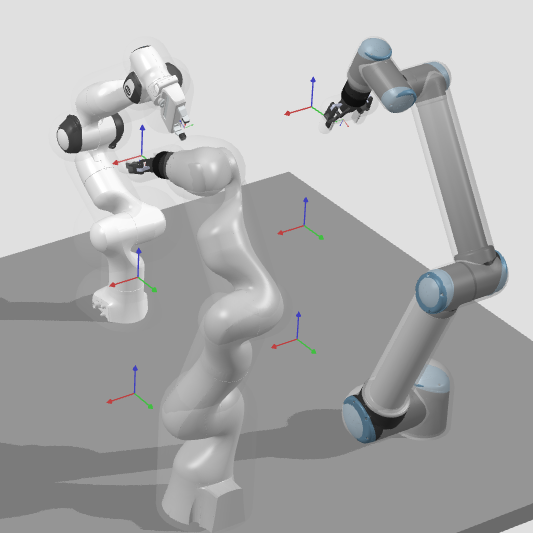}
        \caption{Multi waypoints.}
    \end{subfigure}\hfill
    \begin{subfigure}[t]{.16\textwidth}
        \centering
        \includegraphics[width=.97\linewidth]{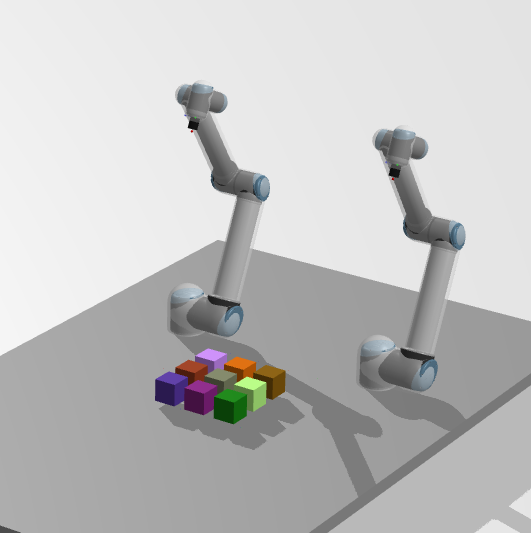}
        \caption{\label{fig:rearrangement}Rearrangement.}
    \end{subfigure}\hfill
    \begin{subfigure}[t]{.16\textwidth}
        \centering
        \includegraphics[width=.97\linewidth]{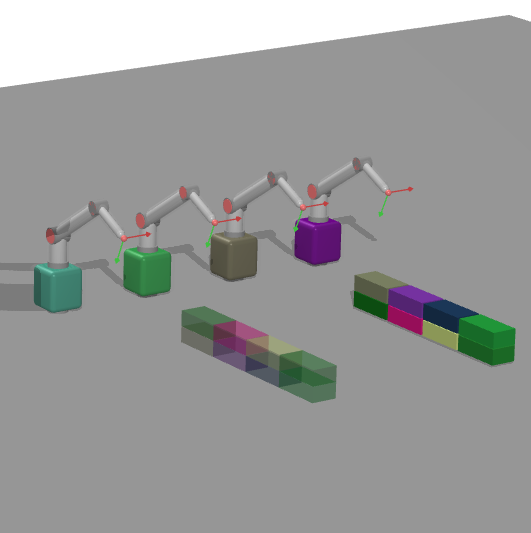}
        \caption{\label{fig:assembly}Mobile assembly.}
    \end{subfigure}
    \caption{The initial states of a selection of problems that are available in the benchmark. The poses that have to be reached by the robots or objects are drawn with lower opacity or indicated with a marker.}
    \label{fig:problems}
\end{figure*}

\subsection{Postprocessing}\label{sec:shortcutting}

When a path is found in any of the planners, we run a postprocessing step to rapidly improve the path cost, which decreases the volume of the informed set. 
Our approach is a minor modification of partial shortcutting \cite{geraerts2007creating}:
The main idea of partial shortcutting is that we do not shortcut all dimensions of the path at once, but select a subset of dimensions from the composite space, and try to shortcut only this subset, while keeping the rest of the plan the same.
Instead of uniformly sampling from all possible subsets, in practice it is more efficient to choose a robot, and then shortcut all dimensions of this robot at once.
Crucially, compared to shortcutting all dimensions at once, this enables improving the unconstrained parts of the path (i.e., the parts that do not need to satisfy goal constraints $g$), while still keeping the simplicity of the shortcutting approach.

\subsection{Proof sketches}
All planners sample uniformly within the modes, and uniformly sample the transitions between modes, therefore gradually expanding the set of sampleable modes, and thus eventually discovering all possible modes (since we assume that the oracle $\mathcal{O}$ gives us \textit{all} possible next task assignments, and thus modes).
Within the modes, all planners work as their respective base versions, and are therefore probabilistically complete, and will eventually connect to the transitions.
If a path exists, the discovered transitions will lead to the terminal mode, which contains the goal pose.
Thus, all planners will find a path if one exists and are therefore probabilistically complete.

The proof sketch for almost-sure asymptotic optimality follows a similar logic:
The search that we run in the PRM* and EIT*/AIT*-planners is optimal, and therefore, as the number of uniform-random samples approaches infinity, we converge to the optimal paths within the modes, and to the optimal mode-transitions. %
Similarly, the rewiring-step in our RRT*-planners is the same as in RRT*, and thus inherits the optimality guarantees from RRT* even over multiple modes.

\section{Experiments}

\begin{figure*}[t]
    \begin{subfigure}[t]{.43\textwidth}
        \centering
        \includegraphics[width=.97\linewidth]{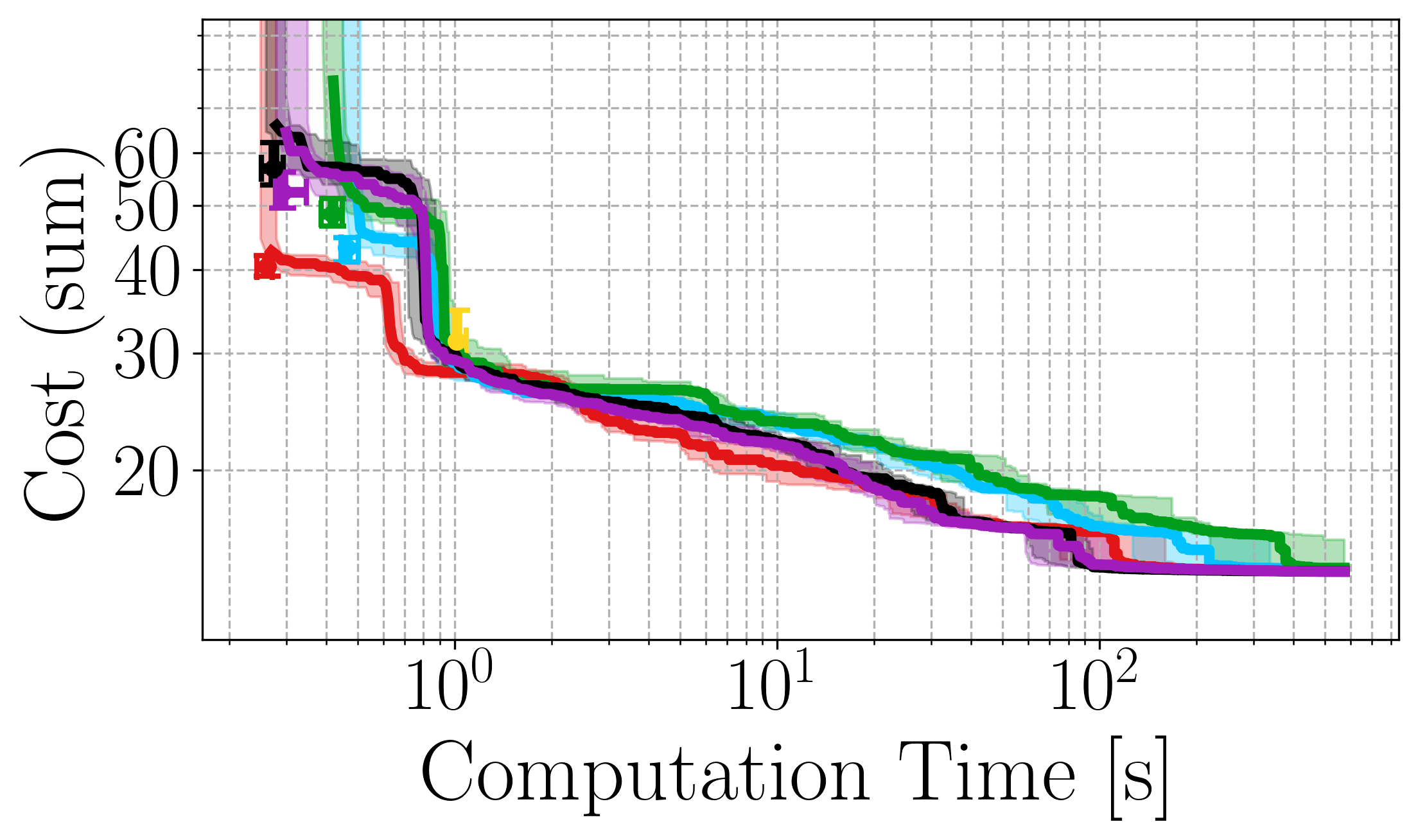}
        \caption{2D hallway (*).}
    \end{subfigure}\hfill
    \begin{subfigure}[t]{.43\textwidth}
        \centering
        \includegraphics[width=.97\linewidth]{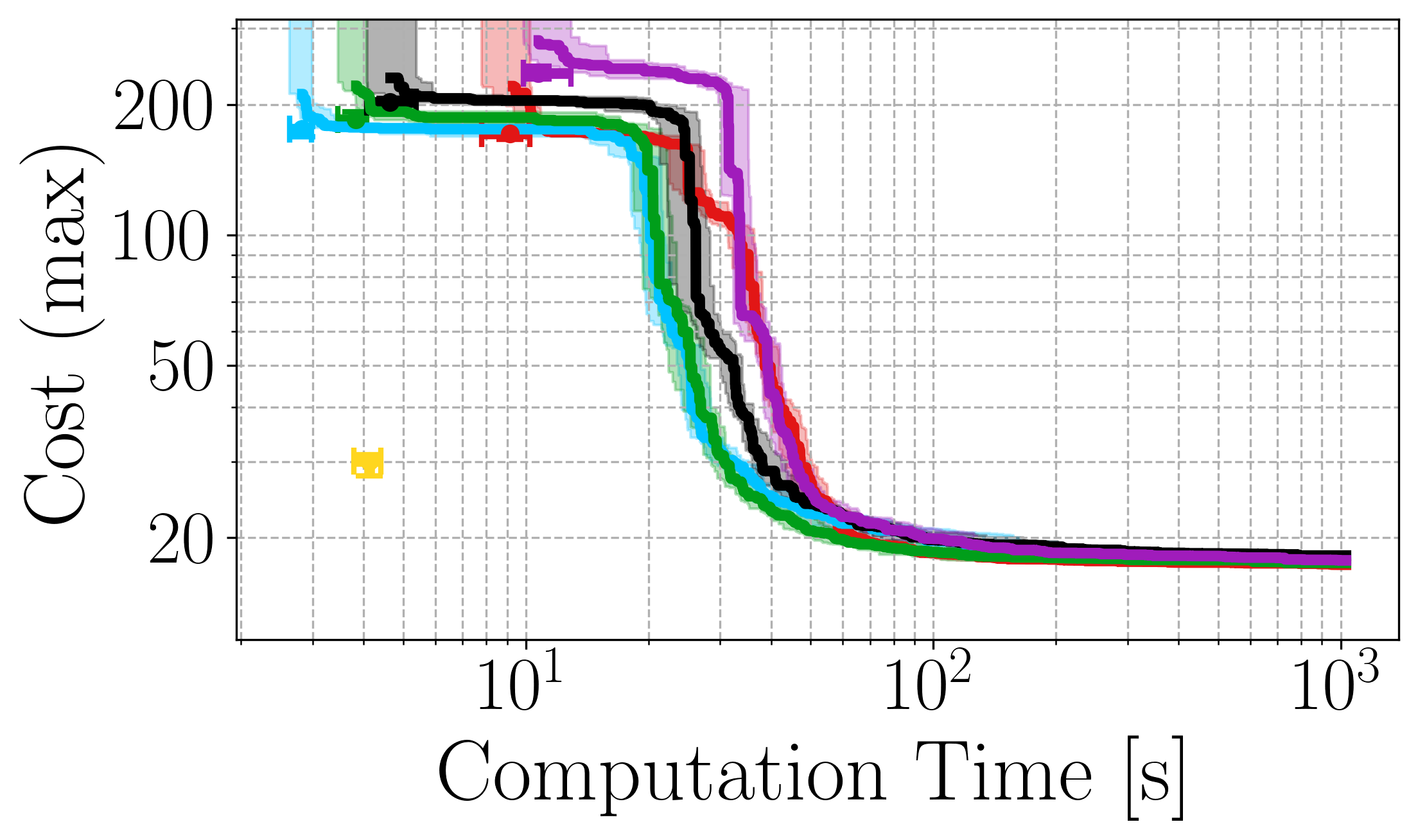}
        \caption{4-arm-box stacking (*).}
    \end{subfigure}\hfill
        \begin{subfigure}[t]{.43\textwidth}
        \centering
        \includegraphics[width=.97\linewidth]{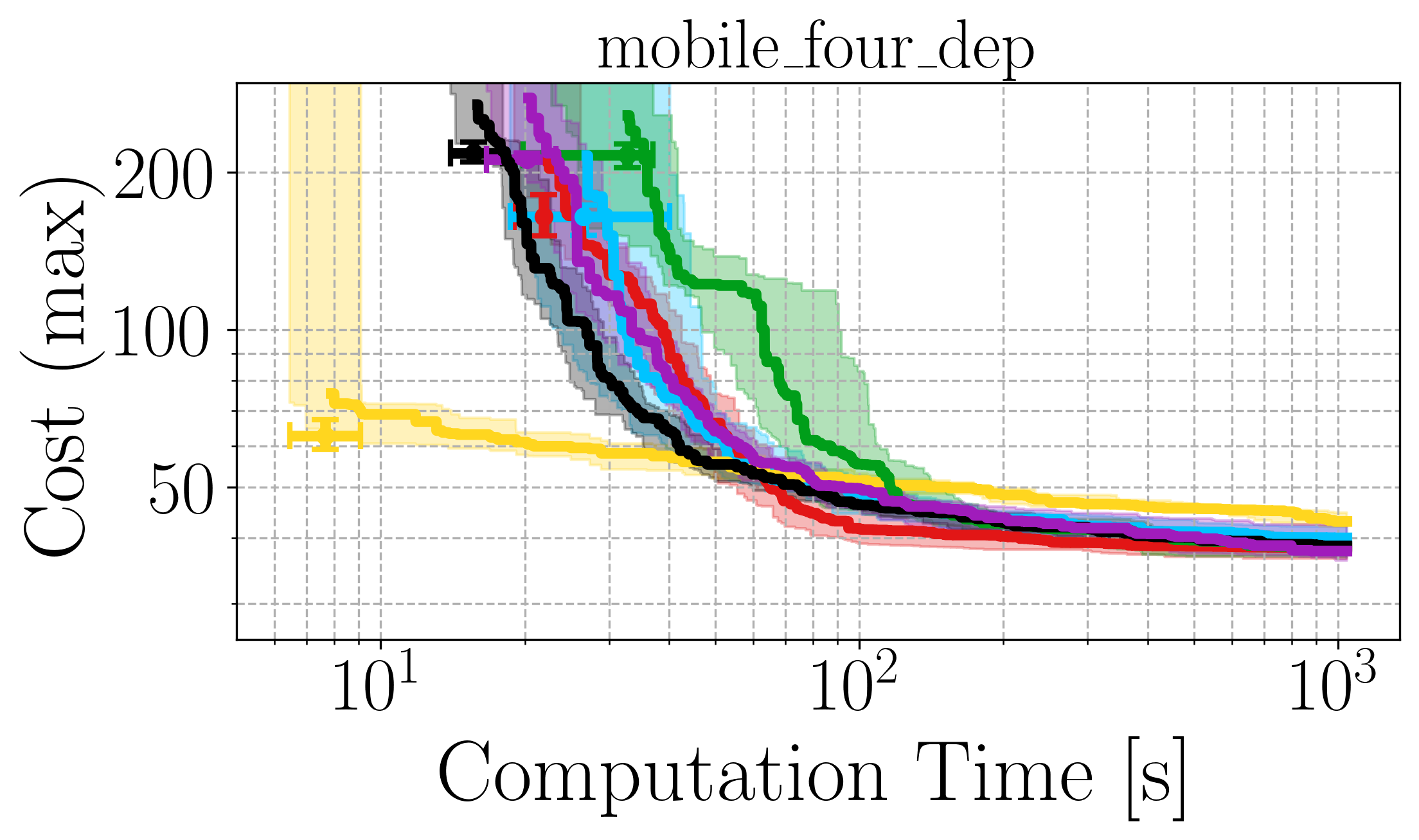}
        \caption{Mobile assembly (**).}
    \end{subfigure}\hfill
    \begin{subfigure}[t]{.43\textwidth}
        \centering
        \includegraphics[width=.97\linewidth]{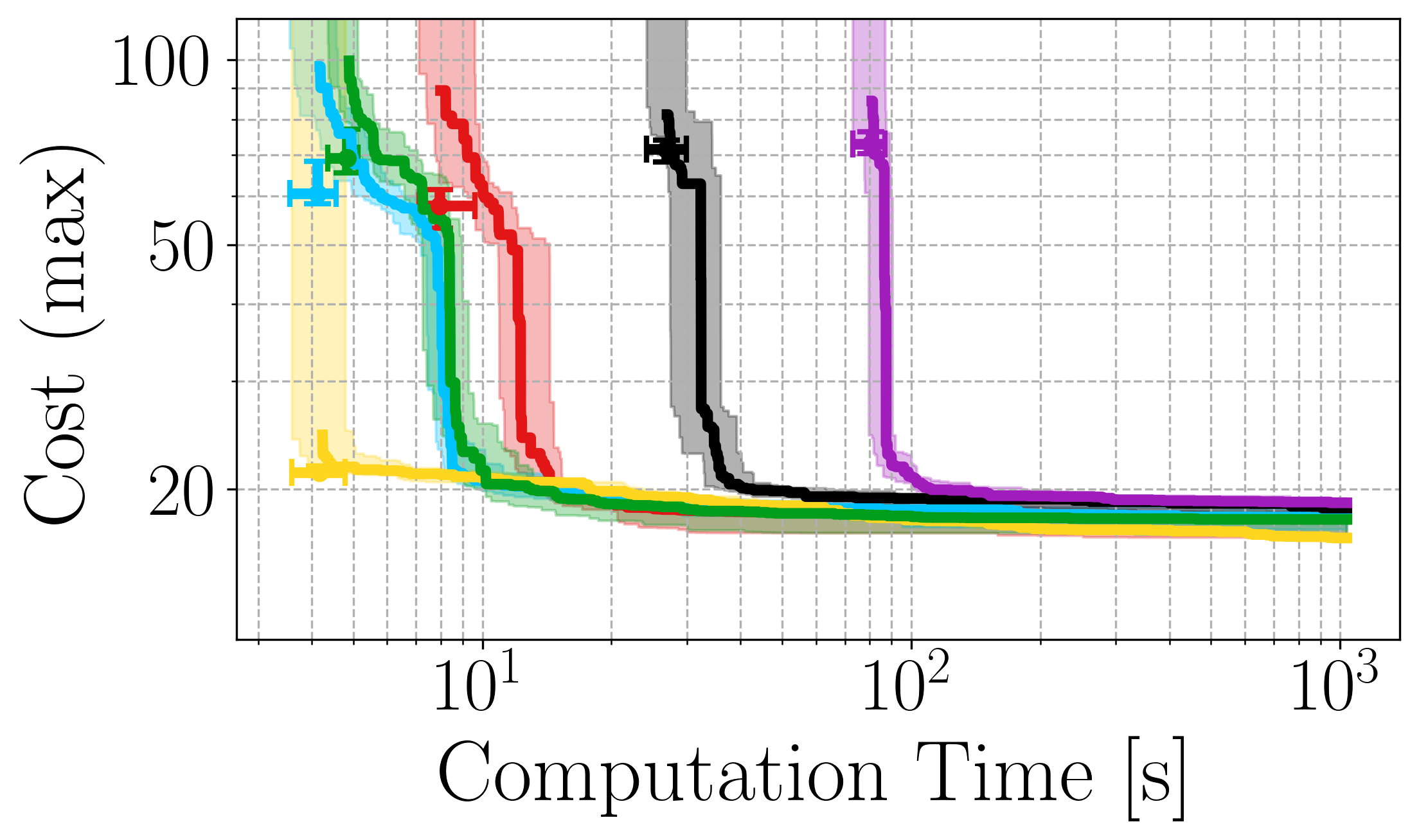}
        \caption{2-arm rearrangement (***).}
    \end{subfigure}
    \begin{subfigure}[b]{1.0\linewidth}%
        \centering
        \begin{tikzpicture}
\begin{axis} [
  width=\textwidth,
  height=0.5\textwidth,
  unbounded coords=jump,
  xtick align=inside,
  ytick align=inside,
  anchor=north,
  hide axis,
  xmajorgrids,
  ymajorgrids,
  major grid style={densely dotted, black!20},
  xmin=0,
  xmax=10,
  ymin=0,
  ymax=10,
  xlabel style={font=\footnotesize},
  xticklabel style={font=\footnotesize},
  ylabel style={font=\footnotesize},
  yticklabel style={font=\footnotesize},
  legend style={anchor=south, legend cell align=left, legend columns=6, at={(axis cs:5, 6)}, font=\small}
]
\addlegendimage{rrt, line width = 1.0pt, mark size=1.0pt, mark=square*}
\addlegendentry{RRT*}

\addlegendimage{birrt, line width = 1.0pt, mark size=1.0pt, mark=square*}
\addlegendentry{BiRRT*}

\addlegendimage{prm, line width = 1.0pt, mark size=1.0pt, mark=square*}
\addlegendentry{PRM*}

\addlegendimage{ait, line width = 1.0pt, mark size=1.0pt, mark=square*}
\addlegendentry{AIT*}

\addlegendimage{eit, line width = 1.0pt, mark size=1.0pt, mark=square*}
\addlegendentry{EIT*}

\addlegendimage{prioritized, line width = 1.0pt, mark size=1.0pt, mark=square*}
\addlegendentry{Prioritized}

\end{axis}
\end{tikzpicture}
    \end{subfigure}
    \caption{Evolution of median cost over time along with the 95\% non-parametric confidence intervals over 50 runs. We also show the median initial solution time and cost using the square with error bars. (*) indicates a fully given task sequence, (**) indicates a partial task ordering, (***) indicates unassigned and unordered tasks. Note that the prioritized planner only acts as anytime planner and improves its cost in the (**) and (***) settings, where multiple different sequences can be generated and planned for.}
    \label{fig:cost_plots}
\end{figure*}

The main contribution of this work is the formalization of the multi-modal, multi-robot, multi-goal planning problem and the adaptation of standard sampling-based planners to the formulation.
We provide implementations of the planners in Python, while the computationally expensive parts, i.e., collision checking and forward kinematics use a more performant backend, which can be easily replaced, allowing for, e.g., GPU parallelization if the backend supports it.
In addition to implementations of the planners, we provide a wide range of benchmarks for the multi-modal, multi-robot, multi-goal planning problem.
The scenarios in our benchmark range from simple settings where the optimal solution path is known to help validate properties of the planner, to more complex scenarios with up to 5 robots (with a total of 30 degrees of freedom).
For many of the problems, there are versions with a complete task ordering and assignment, and others that are either unordered, partially ordered, or where tasks are unassigned.
We show a selection of scenarios in \cref{fig:problems}.

\subsection{Experiments}
We present a selection of experiments to showcase the different types of scenarios and task sequence specifications and compare the optimal planners to a suboptimal prioritized planner \cite{23-hartmann-robplan} on them\footnote{Synchronized planners and standard MAPF planners are not applicable to these problems, and are thus not part of the benchmark.}.
We show four scenarios: (a) A 2D scenario with two robots that is similar to the classic wall-gap, where the robots have to switch positions and go back to their start poses (6D conf. space, 3 subgoals, \cref{fig:hallway}), (b) a scenario with 4 robot arms where 8 boxes have to be stacked (24D conf. space, 17 subgoals, \cref{fig:intro_image}), (c) a scenario involving 4 mobile manipulators rearranging a wall (24D conf. space, 17 subgoals, \cref{fig:assembly}), and (d) a scenario with 2 robot arms that have to collaborate with handovers to rearrange and reorient 9 boxes (12D conf. space, 28 subgoals, \cref{fig:rearrangement}).
In these scenarios, the task sequence is fully determined for scenarios (a) and (b), \textit{partially given} for scenario (c) and \textit{not assigned or ordered} for scenario (d).
To deal with partially ordered problems in the prioritized planner, we generate valid random sequences, and plan for those until the time runs out.

We use the cost function with $w=1$ (i.e., a path-length cost) in scenario (a), and $w=0.01$ (i.e., a min-makespan cost with a small path-length regularization) in scenarios (b)-(d).
We report the median solution costs over 50 runs with different random seeds.
We use a frontier mode sampling strategy ($p=0.98, \epsilon=0$) until a path is found, and then switch to uniform mode sampling in all planners if not stated otherwise. 

The experiments were run with Python 3.10 on a Ryzen 7 5800X (8-core, 4'491 MHz) and 32 GB RAM.

\begin{figure}[t]
    \centering

    \includegraphics[width=0.22\textwidth]{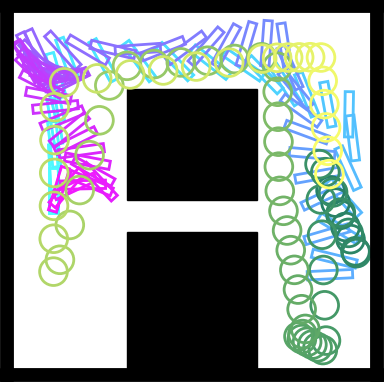}\hfill
    \includegraphics[width=0.22\textwidth]{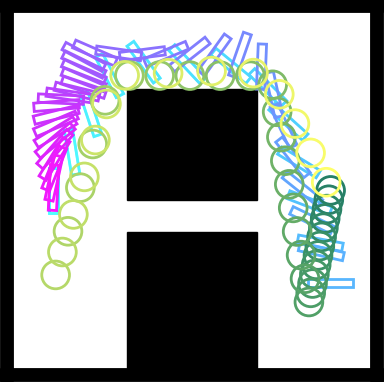}\hfill
    \includegraphics[width=0.22\textwidth]{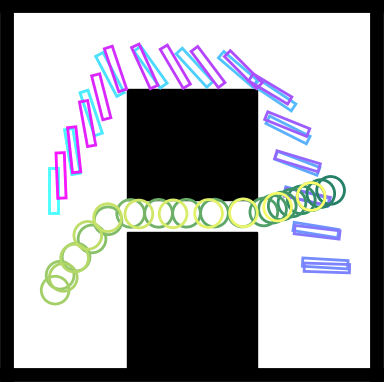}\hfill
    \includegraphics[width=0.22\textwidth]{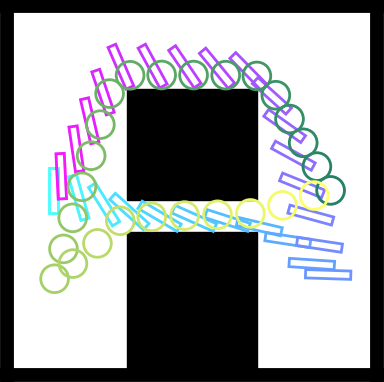}

    \caption{The path at different iterations in the process, from the initial to the optimized solution. Color progression indicates time from start to end (rectangle  
    \smash{
    \raisebox{0.1ex}{%
    \begin{tikzpicture}
        \begin{axis}[
            hide axis,
            scale only axis,
            height=0pt,
            width=1cm,
            colormap={pinktoblue}{
                rgb255(0cm)=(255,20,147);
                rgb255(1cm)=(173,136,206);
                rgb255(2cm)=(102,170,230);
                rgb255(3cm)=(0,0,255);
            },
            colorbar horizontal,
            point meta min=-1,
            point meta max=1,
            colorbar style={
                width=1.5cm,
                height=1.2ex,
                xtick=\empty,
                enlargelimits=false,
                draw opacity=0,
                tickwidth=0pt,
                tick style={draw=none},
            }]
        \end{axis}
    \end{tikzpicture}%
    }
    }
    , disk    
    \smash{\raisebox{0.1ex}{%
    \begin{tikzpicture}
        \begin{axis}[
            hide axis,
            scale only axis,
            height=0pt,
            width=1cm,
            colormap={yellowtodarkgreen}{
                rgb255(0cm)=(255,255,0);
                rgb255(1.5cm)=(58,250,137);
                rgb255(3cm)=(3,141,141);
            },
            colorbar horizontal,
            point meta min=-1,
            point meta max=1,
            colorbar style={
                width=1.5cm,
                height=1.2ex,
                xtick=\empty,
                enlargelimits=false,
                draw opacity=0,
                tickwidth=0pt,
                tick style={draw=none},
            }]
        \end{axis}
    \end{tikzpicture}%
    }}
    )
    .}
    \label{fig:intro}
\end{figure}

\subsection{Results}

\Cref{fig:cost_plots} shows the resulting cost evolution plots.
\arxiv{Success rates over time for the same experiments can be found in the appendix in \cref{fig:success_plots}.}
\Cref{fig:intro} shows the evolution of a path for the hallway problem.
\Cref{tab:all_exp} summarizes a more extensive evaluation of selected planners on a wider range of scenarios. 
First, we want to point out that the prioritized planner does not find solutions on some problems, highlighting again that the prioritized planner is not complete for some problem-settings.

Comparing the optimal planners introduced in this work and the prioritized planner shows that the suboptimal planner with our addition of postprocessing the path finds reasonable solutions: the cost is between 20\% and 70\% over the optimal solution, but better than the initial solutions of the optimal planners.
The optimal planners generally converge to solutions with lower cost, but do so at a later time.
Surprisingly, the planners that plan in composite space find initial solutions faster than the suboptimal planner in many settings.
This is likely because (1) finding a \textit{feasible} path is relatively simple in composite space in many scenarios, where the robots can simply be `moved out of the way' and (2) fixing previously planned paths can restrict the feasible space too much. %

\begin{figure}[t]
    \centering
    \includegraphics[width=0.3\linewidth]{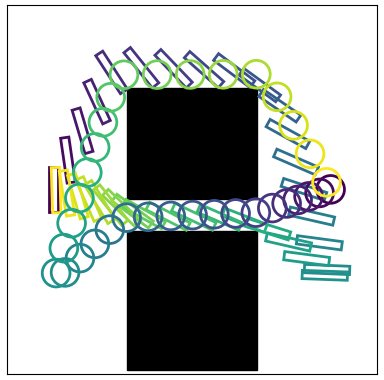}
    \includegraphics[width=0.3\linewidth]{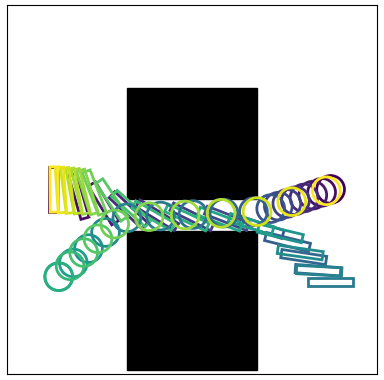}
    \caption{The optimal paths when using a max \textbf{(left)} or sum \textbf{(right)} cost function in the 2D hallway scenario, where the robots have to reach a goal on the other side and return. Color indicates time from purple (start) to yellow (end).}
    \label{fig:sum_vs_max}
\end{figure}

\paragraph{Influence of the cost function}
We illustrate the difference of the sum-cost and the max-cost on the hallway example in \cref{fig:sum_vs_max}:
Both robots go through the wall gap twice when using the sum-cost, and do not make use of the passage at the top, compared to the (regularized) max-cost, where they make use of the passage at the top.

\paragraph{Convergence in complex problems}
We can see in \cref{tab:all_exp} that the optimal planners do not converge to the same cost for some of the problems in the allocated time.
This highlights that BiRRT* has a slower convergence rate compared to AIT*.
We did run a subset of the scenarios for longer in order to confirm that they do indeed converge to the same costs.

In scenarios (c) and (d) and the unordered and unassigned scenarios in \cref{tab:all_exp}, the planners do not reliably converge to the same solution at the end of their runtime but remain 5\%-10\% above the lowest found solution, showing that some of the found solutions are not optimal.
This is because the problem space is extremely large due to the discrete choice of task assignment and ordering in addition to the high dimensionality of the continuous space.
Thus, while the planners are theoretically optimal, in practice, we find that the planners tend to get stuck in a local optimum and do not switch the discrete assignment and ordering often.
Note that in scenario (d), the median final cost of the prioritized planner is lower than the median final cost of the composite space planners.
This is likely due to the better exploration of different sequences.

\subsection{Ablations}
We present ablations to explain the design decisions that deviate from the standard versions of the planners.
We want to emphasize again that while the core ideas of the planners that we use here are the same as in their original versions, the adaptions are crucial in order to obtain the planning times and costs that we show here.
In the following, we do not show the results for all planners as the effects are similar for most, and would crowd the figures.

\begin{figure}[t]
    \centering
    \includegraphics[width=0.43\linewidth]{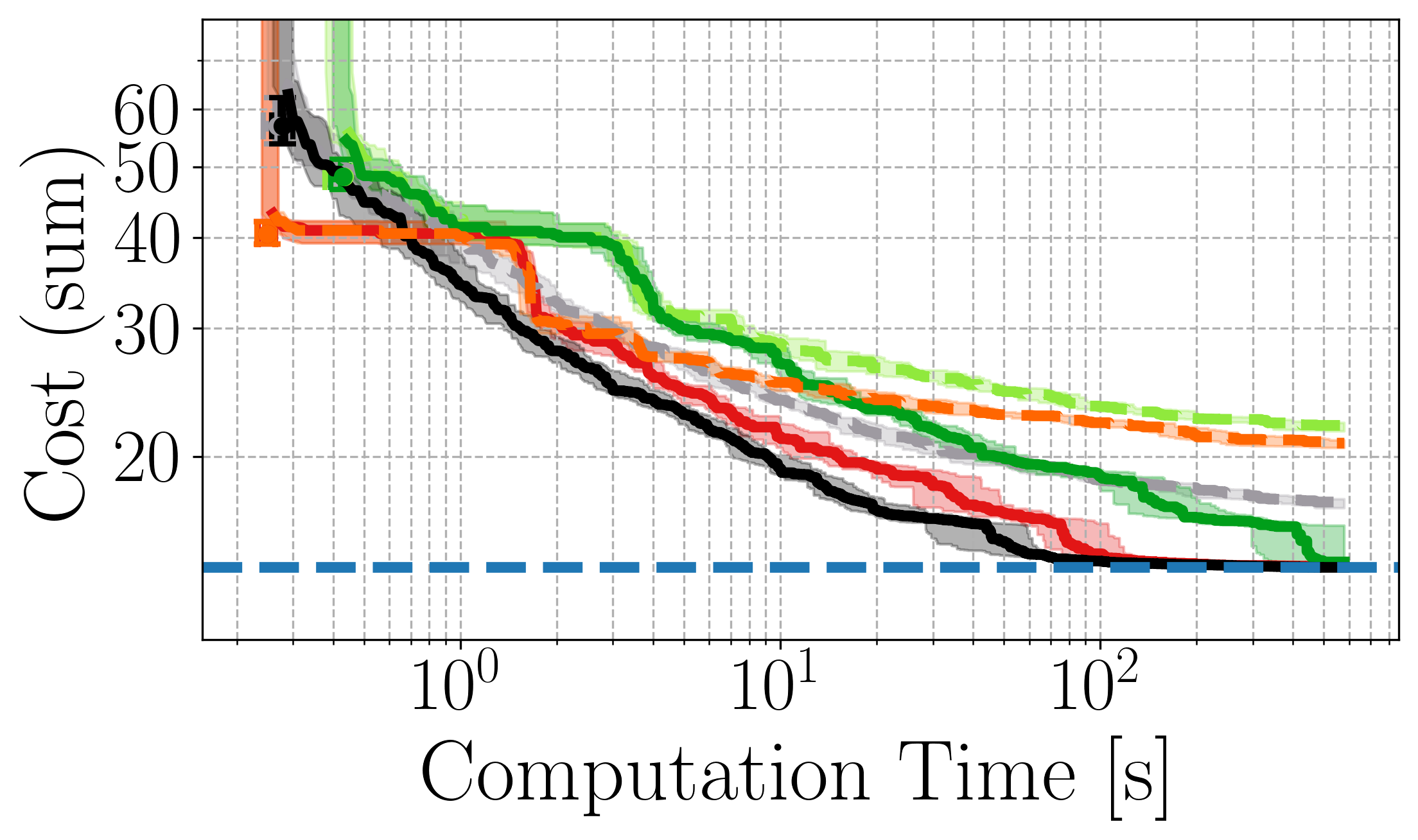}
    \includegraphics[width=0.43\linewidth]{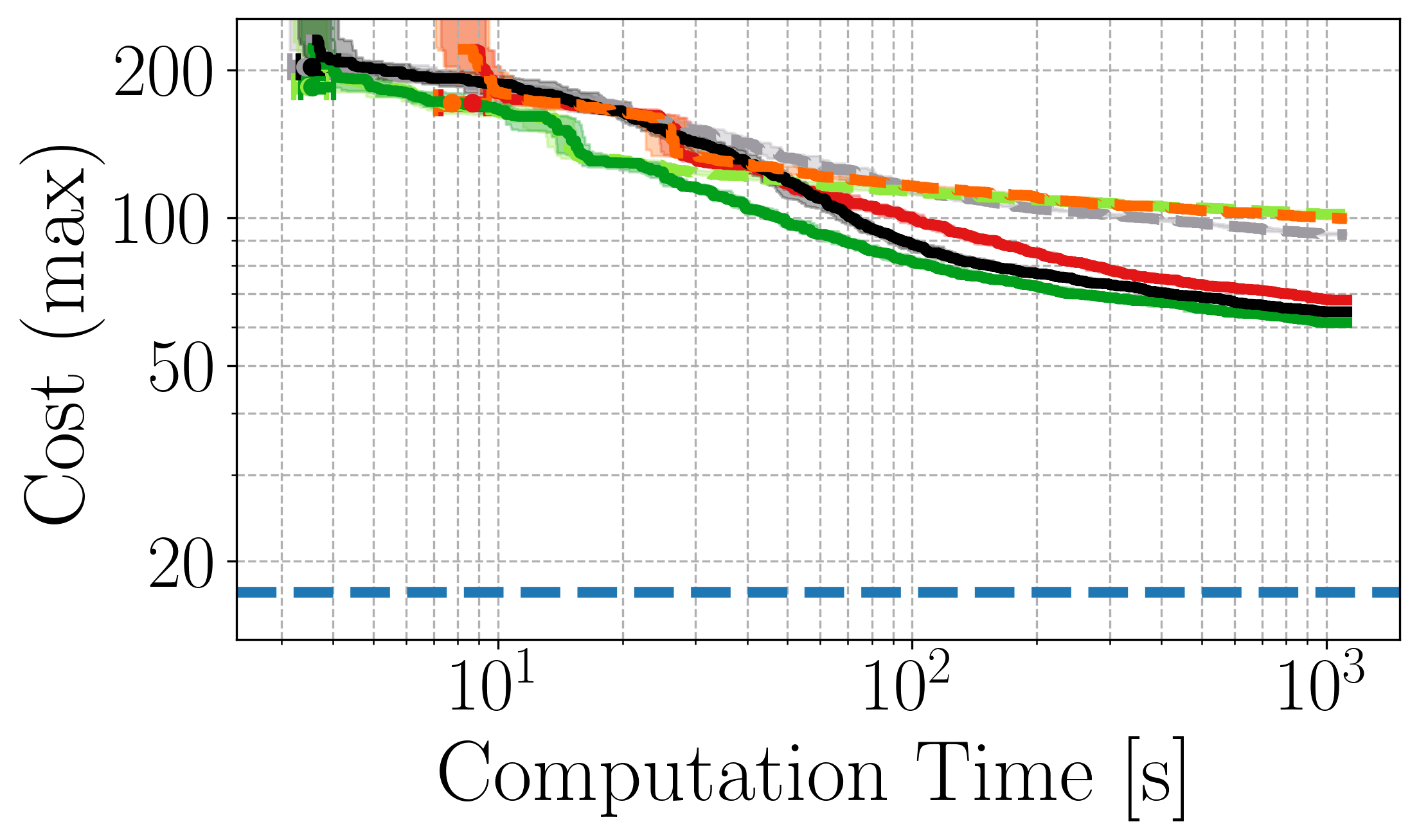}
    \begin{subfigure}[b]{1.0\linewidth}%
        \centering
        \begin{tikzpicture}
\begin{axis} [
  width=\textwidth,
  height=0.5\textwidth,
  unbounded coords=jump,
  xtick align=inside,
  ytick align=inside,
  anchor=north,
  hide axis,
  xmajorgrids,
  ymajorgrids,
  major grid style={densely dotted, black!20},
  xmin=0,
  xmax=10,
  ymin=0,
  ymax=10,
  xlabel style={font=\footnotesize},
  xticklabel style={font=\footnotesize},
  ylabel style={font=\footnotesize},
  yticklabel style={font=\footnotesize},
   legend style={anchor=center, legend cell align=left, legend columns=7, at={(0.5, 0.5)}, font=\small},
]

\addlegendimage{
legend image code/.code={
        \def\linestart{0.18cm}
        \def\lineend{0.62cm}
        \def\centerx{0.4cm}
        \def\squaresize{0.07cm}
        \def\offset{0.05cm}
        \draw[line width=1pt, faded1 birrt] (\linestart,\offset) -- (\lineend,\offset);
        \filldraw[fill=faded1 birrt, draw=faded1 birrt] 
            ({\centerx-0.5*\squaresize},{\offset-0.5*\squaresize}) 
            rectangle 
            ({\centerx+0.5*\squaresize},{\offset+0.5*\squaresize});
        \draw[line width=1pt, birrt] (\linestart,-\offset) -- (\lineend,-\offset);
        \filldraw[fill=birrt, draw=birrt] 
            ({\centerx-0.5*\squaresize},{-\offset-0.5*\squaresize}) 
            rectangle 
            ({\centerx+0.5*\squaresize},{-\offset+0.5*\squaresize});
    }
}
\addlegendentry{BiRRT*}

\addlegendimage{
legend image code/.code={
        \def\linestart{0.18cm}
        \def\lineend{0.62cm}
        \def\centerx{0.4cm}
        \def\squaresize{0.07cm}
        \def\offset{0.05cm}
        \draw[line width=1pt, faded1 prm] (\linestart,\offset) -- (\lineend,\offset);
        \filldraw[fill=faded1 prm, draw=faded1 prm] 
            ({\centerx-0.5*\squaresize},{\offset-0.5*\squaresize}) 
            rectangle 
            ({\centerx+0.5*\squaresize},{\offset+0.5*\squaresize});
        \draw[line width=1pt, prm] (\linestart,-\offset) -- (\lineend,-\offset);
        \filldraw[fill=prm, draw=prm] 
            ({\centerx-0.5*\squaresize},{-\offset-0.5*\squaresize}) 
            rectangle 
            ({\centerx+0.5*\squaresize},{-\offset+0.5*\squaresize});
    }
}
\addlegendentry{PRM*}

\addlegendimage{
legend image code/.code={
        \def\linestart{0.18cm}
        \def\lineend{0.62cm}
        \def\centerx{0.4cm}
        \def\squaresize{0.07cm}
        \def\offset{0.05cm}
        \draw[line width=1pt, faded1 eit] (\linestart,\offset) -- (\lineend,\offset);
        \filldraw[fill=faded1 eit, draw=faded1 eit] 
            ({\centerx-0.5*\squaresize},{\offset-0.5*\squaresize}) 
            rectangle 
            ({\centerx+0.5*\squaresize},{\offset+0.5*\squaresize});
        \draw[line width=1pt, eit] (\linestart,-\offset) -- (\lineend,-\offset);
        \filldraw[fill=eit, draw=eit] 
            ({\centerx-0.5*\squaresize},{-\offset-0.5*\squaresize}) 
            rectangle 
            ({\centerx+0.5*\squaresize},{-\offset+0.5*\squaresize});
    }
}
\addlegendentry{EIT*}

\end{axis}
\end{tikzpicture}
    \end{subfigure}
    \caption{Local (solid) and global (dashed) informed sampling on the 2D hallway \textbf{(left)} and the 4-arm box-stacking \textbf{(right)} scenarios. The dashed blue line is the best cost of the non-ablated planners.}
    \label{fig:informed_sampling}
\end{figure}

\paragraph{Locally informed sampling}
Only using standard informed sampling in the problems we consider here is insufficient, as the informed set is large compared to the configuration space, and thus effectively equivalent to uniform sampling.
In \cref{fig:informed_sampling}, the planners do not use shortcutting in order to better highlight the difference between the ablations.
The planners using locally informed sampling outperform the planners that do use standard informed sampling.

\begin{figure}[t]
    \centering
    \includegraphics[width=0.43\linewidth]{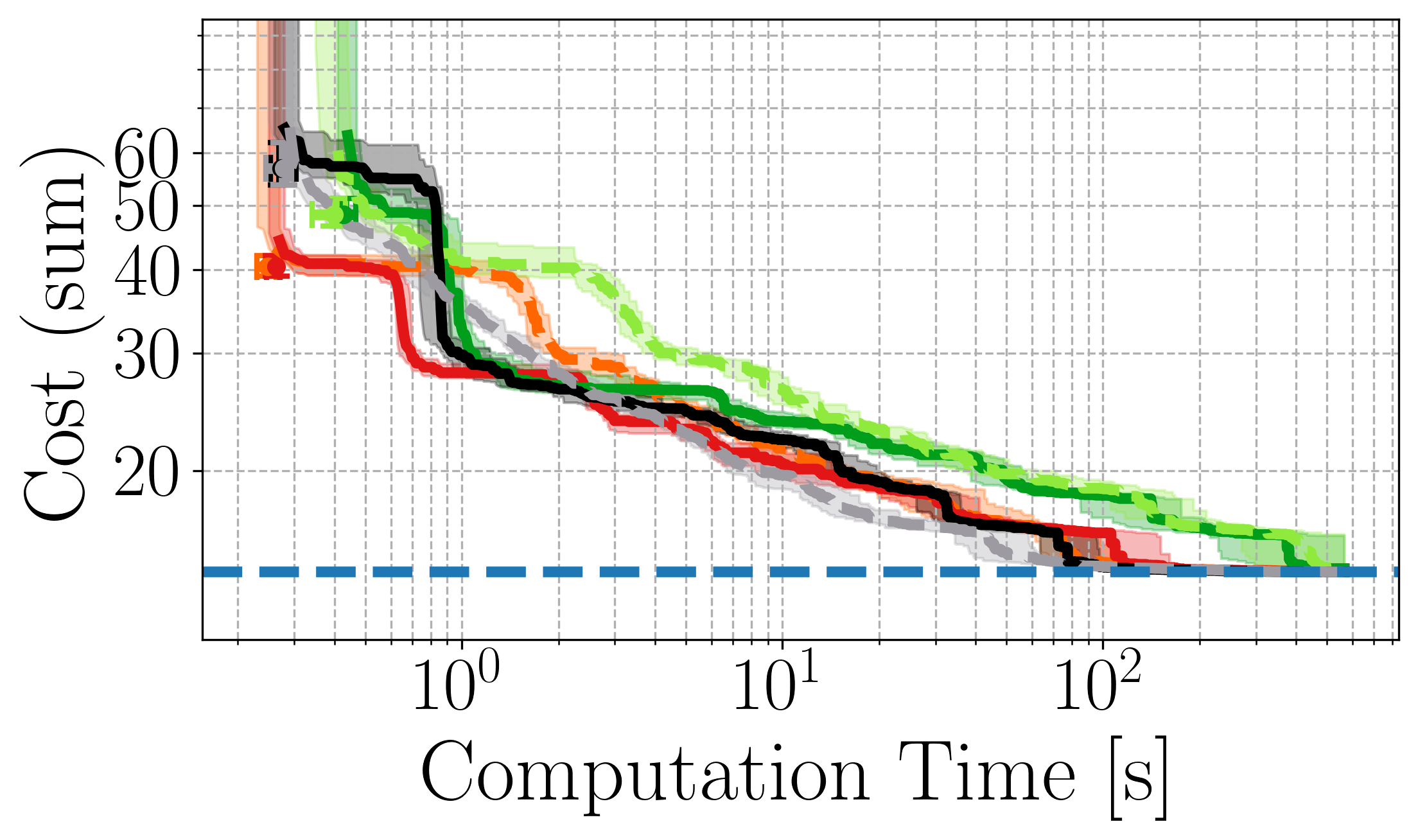}
    \includegraphics[width=0.43\linewidth]{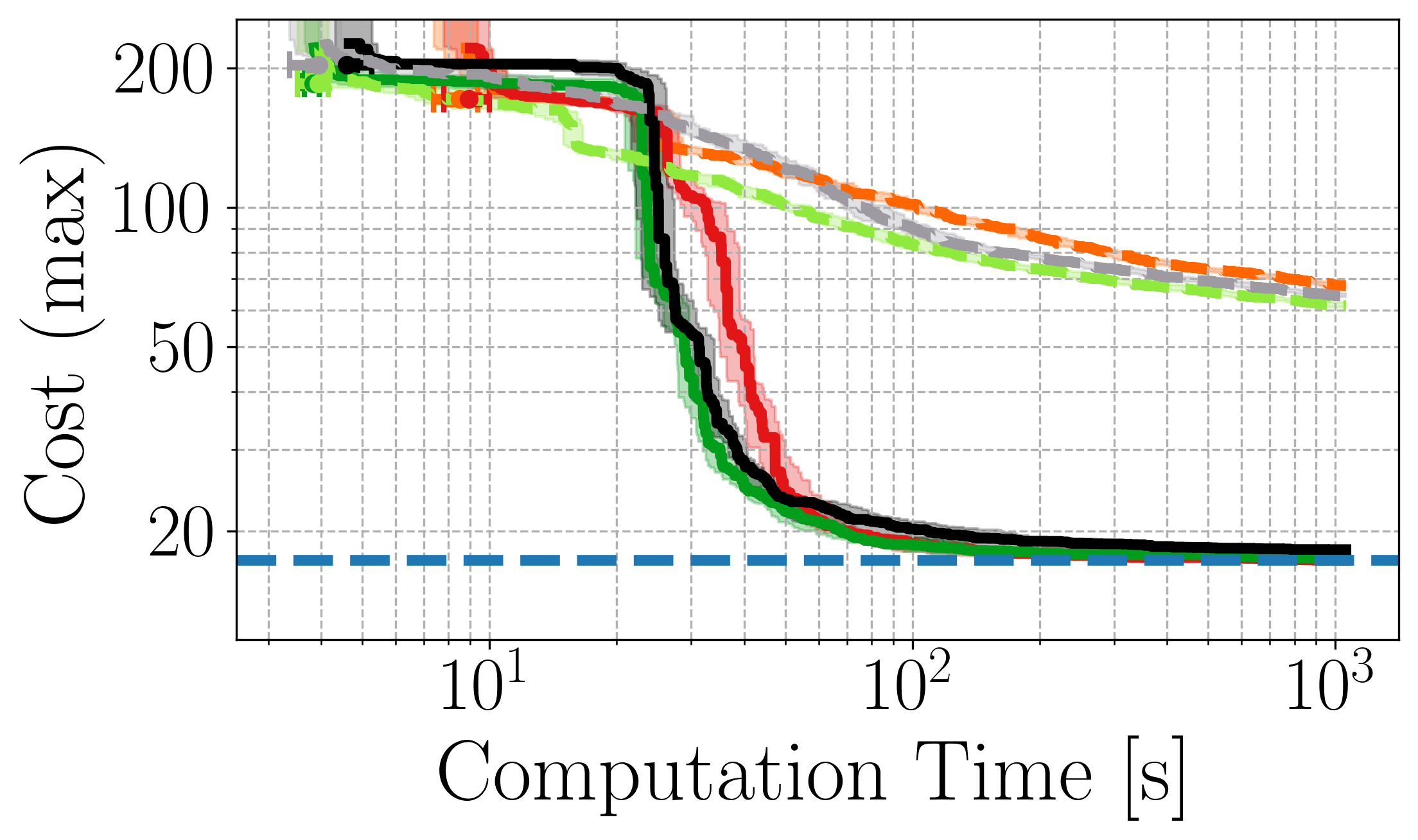}
    \begin{subfigure}[b]{1.0\linewidth}%
        \centering
        \begin{tikzpicture}
\begin{axis} [
  width=\textwidth,
  height=0.5\textwidth,
  unbounded coords=jump,
  xtick align=inside,
  ytick align=inside,
  anchor=north,
  hide axis,
  xmajorgrids,
  ymajorgrids,
  major grid style={densely dotted, black!20},
  xmin=0,
  xmax=10,
  ymin=0,
  ymax=10,
  xlabel style={font=\footnotesize},
  xticklabel style={font=\footnotesize},
  ylabel style={font=\footnotesize},
  yticklabel style={font=\footnotesize},
   legend style={anchor=center, legend cell align=left, legend columns=7, at={(0.5, 0.5)}, font=\small},
]

\addlegendimage{
legend image code/.code={
        \def\linestart{0.18cm}
        \def\lineend{0.62cm}
        \def\centerx{0.4cm}
        \def\squaresize{0.07cm}
        \def\offset{0.05cm}
        \draw[line width=1pt, faded1 birrt] (\linestart,\offset) -- (\lineend,\offset);
        \filldraw[fill=faded1 birrt, draw=faded1 birrt] 
            ({\centerx-0.5*\squaresize},{\offset-0.5*\squaresize}) 
            rectangle 
            ({\centerx+0.5*\squaresize},{\offset+0.5*\squaresize});
        \draw[line width=1pt, birrt] (\linestart,-\offset) -- (\lineend,-\offset);
        \filldraw[fill=birrt, draw=birrt] 
            ({\centerx-0.5*\squaresize},{-\offset-0.5*\squaresize}) 
            rectangle 
            ({\centerx+0.5*\squaresize},{-\offset+0.5*\squaresize});
    }
}
\addlegendentry{BiRRT*}

\addlegendimage{
legend image code/.code={
        \def\linestart{0.18cm}
        \def\lineend{0.62cm}
        \def\centerx{0.4cm}
        \def\squaresize{0.07cm}
        \def\offset{0.05cm}
        \draw[line width=1pt, faded1 prm] (\linestart,\offset) -- (\lineend,\offset);
        \filldraw[fill=faded1 prm, draw=faded1 prm] 
            ({\centerx-0.5*\squaresize},{\offset-0.5*\squaresize}) 
            rectangle 
            ({\centerx+0.5*\squaresize},{\offset+0.5*\squaresize});
        \draw[line width=1pt, prm] (\linestart,-\offset) -- (\lineend,-\offset);
        \filldraw[fill=prm, draw=prm] 
            ({\centerx-0.5*\squaresize},{-\offset-0.5*\squaresize}) 
            rectangle 
            ({\centerx+0.5*\squaresize},{-\offset+0.5*\squaresize});
    }
}
\addlegendentry{PRM*}

\addlegendimage{
legend image code/.code={
        \def\linestart{0.18cm}
        \def\lineend{0.62cm}
        \def\centerx{0.4cm}
        \def\squaresize{0.07cm}
        \def\offset{0.05cm}
        \draw[line width=1pt, faded1 eit] (\linestart,\offset) -- (\lineend,\offset);
        \filldraw[fill=faded1 eit, draw=faded1 eit] 
            ({\centerx-0.5*\squaresize},{\offset-0.5*\squaresize}) 
            rectangle 
            ({\centerx+0.5*\squaresize},{\offset+0.5*\squaresize});
        \draw[line width=1pt, eit] (\linestart,-\offset) -- (\lineend,-\offset);
        \filldraw[fill=eit, draw=eit] 
            ({\centerx-0.5*\squaresize},{-\offset-0.5*\squaresize}) 
            rectangle 
            ({\centerx+0.5*\squaresize},{-\offset+0.5*\squaresize});
    }
}
\addlegendentry{EIT*}

\end{axis}
\end{tikzpicture}
    \end{subfigure}
    \caption{Planners with (solid) and without (dashed) shortcutting on the 2D hallway \textbf{(left)} and the 4-arm box-stacking \textbf{(right)} scenarios. All planners are using locally informed sampling. The dashed blue line is the best cost of the non-ablated planners.}
    \label{fig:shortcutting}
\end{figure}

\paragraph{Shortcutting in the planner}
Shortcutting a path after it was found leads to faster convergence than using only informed sampling (\cref{fig:shortcutting}) while not affecting the optimality or completeness properties of the planners.
This effect is more pronounced in high-dimensional scenarios.

\begin{figure}[t]
    \centering
    \includegraphics[width=0.43\linewidth]{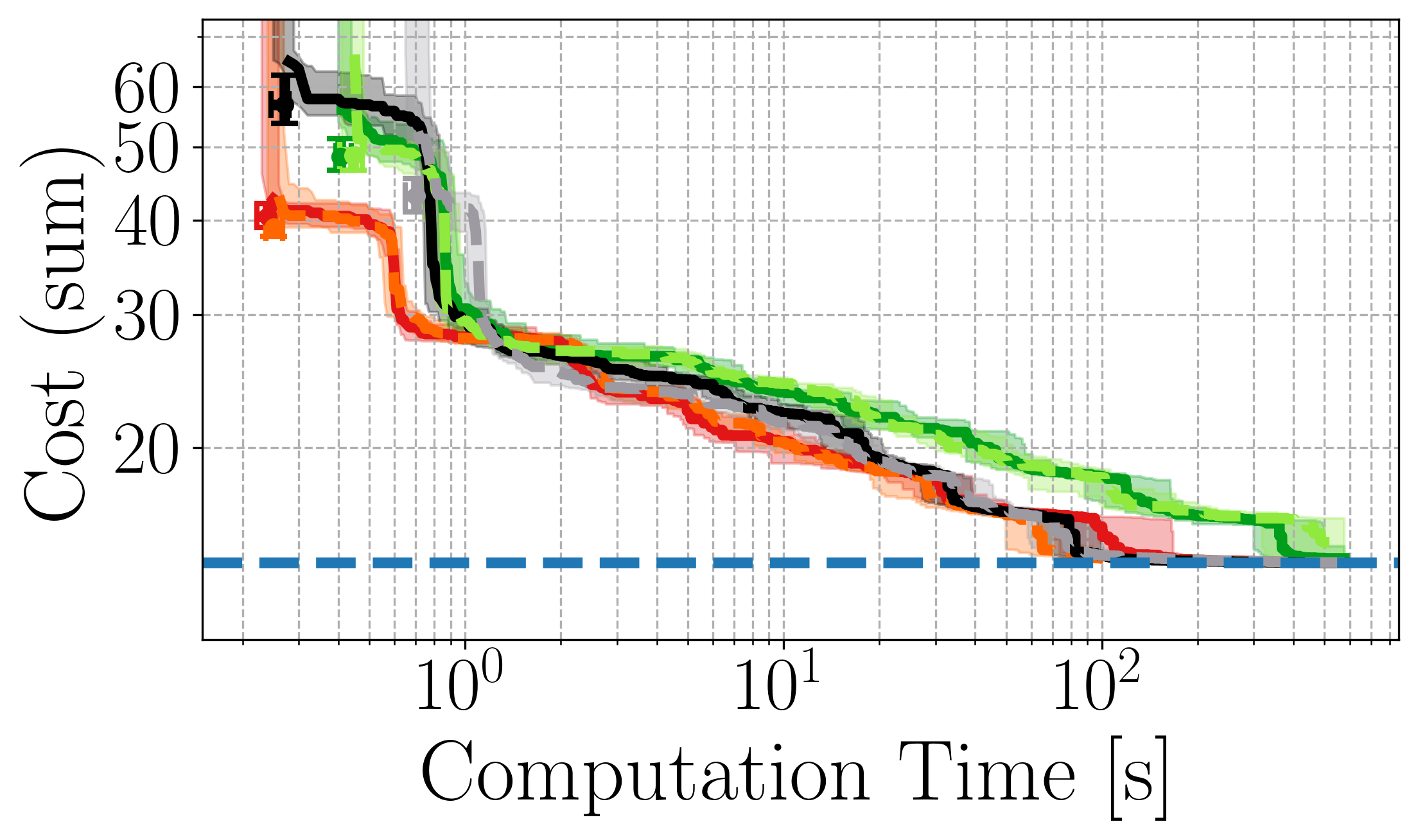}
    \includegraphics[width=0.43\linewidth]{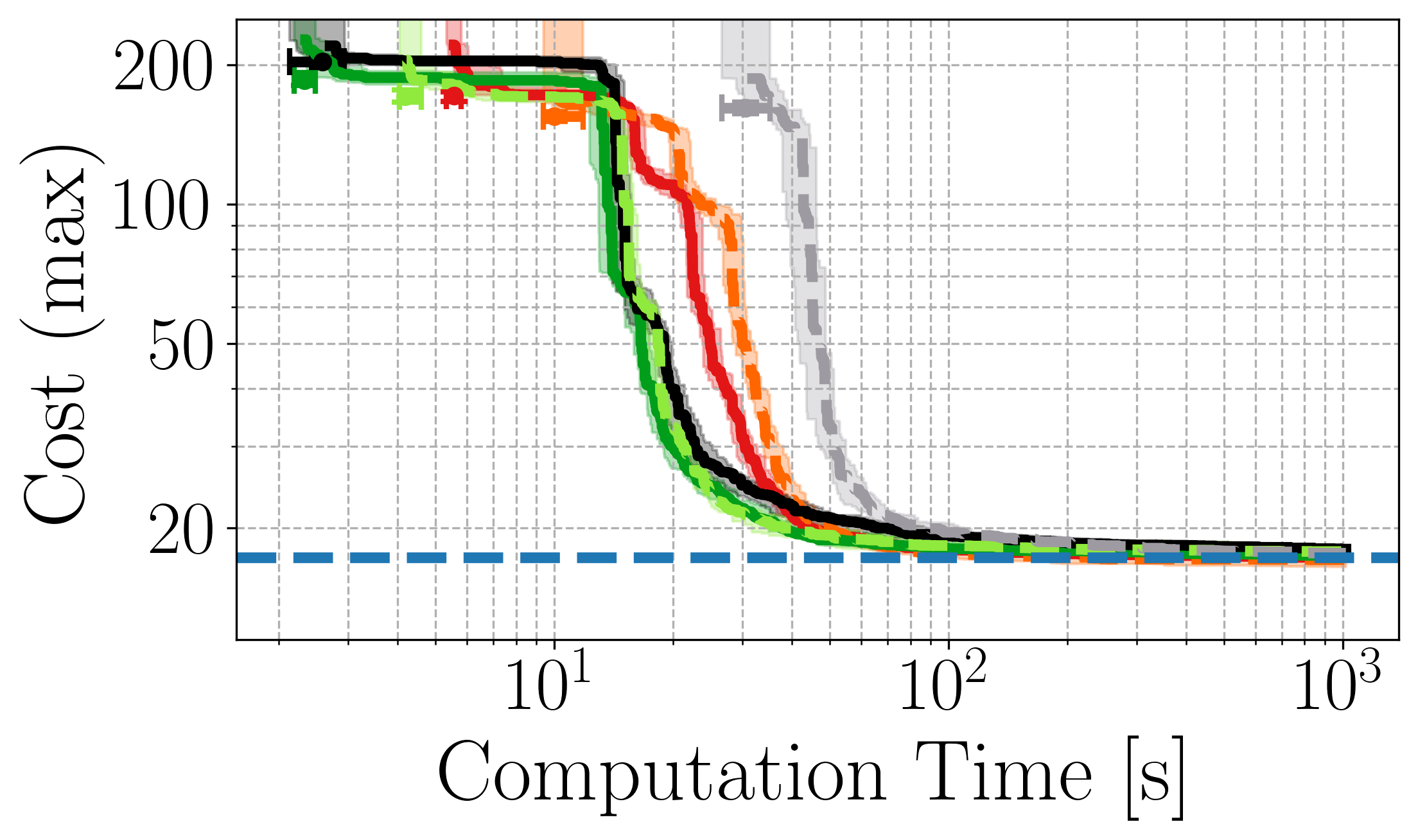}
    \par\medskip
    \begin{minipage}{\linewidth}
        \centering
        \begin{tikzpicture}
\begin{axis} [
  width=\textwidth,
  height=0.5\textwidth,
  unbounded coords=jump,
  xtick align=inside,
  ytick align=inside,
  anchor=north,
  hide axis,
  xmajorgrids,
  ymajorgrids,
  major grid style={densely dotted, black!20},
  xmin=0,
  xmax=10,
  ymin=0,
  ymax=10,
  xlabel style={font=\footnotesize},
  xticklabel style={font=\footnotesize},
  ylabel style={font=\footnotesize},
  yticklabel style={font=\footnotesize},
   legend style={anchor=center, legend cell align=left, legend columns=7, at={(0.5, 0.5)}, font=\small},
]

\addlegendimage{
legend image code/.code={
        \def\linestart{0.18cm}
        \def\lineend{0.62cm}
        \def\centerx{0.4cm}
        \def\squaresize{0.07cm}
        \def\offset{0.05cm}
        \draw[line width=1pt, faded1 birrt] (\linestart,\offset) -- (\lineend,\offset);
        \filldraw[fill=faded1 birrt, draw=faded1 birrt] 
            ({\centerx-0.5*\squaresize},{\offset-0.5*\squaresize}) 
            rectangle 
            ({\centerx+0.5*\squaresize},{\offset+0.5*\squaresize});
        \draw[line width=1pt, birrt] (\linestart,-\offset) -- (\lineend,-\offset);
        \filldraw[fill=birrt, draw=birrt] 
            ({\centerx-0.5*\squaresize},{-\offset-0.5*\squaresize}) 
            rectangle 
            ({\centerx+0.5*\squaresize},{-\offset+0.5*\squaresize});
    }
}
\addlegendentry{BiRRT*}

\addlegendimage{
legend image code/.code={
        \def\linestart{0.18cm}
        \def\lineend{0.62cm}
        \def\centerx{0.4cm}
        \def\squaresize{0.07cm}
        \def\offset{0.05cm}
        \draw[line width=1pt, faded1 prm] (\linestart,\offset) -- (\lineend,\offset);
        \filldraw[fill=faded1 prm, draw=faded1 prm] 
            ({\centerx-0.5*\squaresize},{\offset-0.5*\squaresize}) 
            rectangle 
            ({\centerx+0.5*\squaresize},{\offset+0.5*\squaresize});
        \draw[line width=1pt, prm] (\linestart,-\offset) -- (\lineend,-\offset);
        \filldraw[fill=prm, draw=prm] 
            ({\centerx-0.5*\squaresize},{-\offset-0.5*\squaresize}) 
            rectangle 
            ({\centerx+0.5*\squaresize},{-\offset+0.5*\squaresize});
    }
}
\addlegendentry{PRM*}

\addlegendimage{
legend image code/.code={
        \def\linestart{0.18cm}
        \def\lineend{0.62cm}
        \def\centerx{0.4cm}
        \def\squaresize{0.07cm}
        \def\offset{0.05cm}
        \draw[line width=1pt, faded1 eit] (\linestart,\offset) -- (\lineend,\offset);
        \filldraw[fill=faded1 eit, draw=faded1 eit] 
            ({\centerx-0.5*\squaresize},{\offset-0.5*\squaresize}) 
            rectangle 
            ({\centerx+0.5*\squaresize},{\offset+0.5*\squaresize});
        \draw[line width=1pt, eit] (\linestart,-\offset) -- (\lineend,-\offset);
        \filldraw[fill=eit, draw=eit] 
            ({\centerx-0.5*\squaresize},{-\offset-0.5*\squaresize}) 
            rectangle 
            ({\centerx+0.5*\squaresize},{-\offset+0.5*\squaresize});
    }
}
\addlegendentry{EIT*}

\end{axis}
\end{tikzpicture}
    \end{minipage}
    \caption{Frontier (solid) and uniform (dashed) initial mode sampling strategies on the 2D hallway \textbf{(left)} and 4-arm box-stacking \textbf{(right)} scenarios. The dashed blue line is the best cost of the non-ablated planners.}
    \label{fig:mode_sampling}
\end{figure}

\paragraph{Mode sampling strategy}
Employing a \textit{frontier sampling strategy} during the initial search significantly accelerates the discovery of a first solution compared to uniform sampling (\cref{fig:mode_sampling}) in high dimensional, long horizon scenarios.
While not shown here, a greedy sampling strategy gives a further significant speedup for the initial solution but does sacrifice completeness if not done carefully.

\begin{table*}
\setlength{\tabcolsep}{3pt}

\caption{Results for more scenarios from the benchmark problems. We show the median of all the statistics of 25 runs. The cost $c_{t_\text{max}}$ is the cost after the time specified.
In addition to the solution of the standard prioritized planner, we also shortcut the plan from the prioritized planner using our shortcutting method, and show the cost under $c_{t_\text{max}}$.}
\label{tab:all_exp}
\tiny
\centering
\begin{tabular}{l|rrr|rrr|rrr|rrr}
\toprule
 & \#Dim & \#Tasks & $t_\text{{max}}$ [s] & \multicolumn{3}{c}{$t_\text{{init}}$ [s]} & \multicolumn{3}{c}{$c_\text{{init}}$} & \multicolumn{3}{c}{$c_{{t_\text{{max}}}}$}  \\
  &  &  & & \multicolumn{1}{c}{Prio}& \multicolumn{1}{c}{BiRRT*}& \multicolumn{1}{c}{AIT*}& \multicolumn{1}{c}{Prio}& \multicolumn{1}{c}{BiRRT*}& \multicolumn{1}{c}{AIT*}& \multicolumn{1}{c}{Prio}& \multicolumn{1}{c}{BiRRT*}& \multicolumn{1}{c}{AIT*} \\
\midrule
simple & $6$ & $3$ & $100$ & $1.09$ & $\mathbf{ 0.17 }$& $0.20$ & $\mathbf{ 4.88 }$& $11.94$ & $6.24$ & $3.69$ & $\mathbf{ 2.27 }$& $2.28$ \\ 
single\_agent\_mover & $3$ & $3$ & $100$ & $-$ & $\mathbf{ 0.09 }$& $0.11$ & $-$ & $8.83$ & $\mathbf{ 4.47 }$& $-$ & $3.15$ & $\mathbf{ 2.90 }$\\ 
2d\_handover & $6$ & $6$ & $500$ & $-$ & $\mathbf{ 4.13 }$& $20.43$ & $-$ & $56.97$ & $\mathbf{ 53.74 }$& $-$ & $\mathbf{ 17.90 }$& $19.21$ \\ 
random\_2d & $9$ & $13$ & $500$ & $-$ & $\mathbf{ 7.95 }$& $31.29$ & $-$ & $124.49$ & $\mathbf{ 90.56 }$& $-$ & $\mathbf{ 26.09 }$& $27.11$ \\ 
other\_hallway & $6$ & $3$ & $500$ & $2.92$ & $0.37$ & $\mathbf{ 0.37 }$& $29.41$ & $35.66$ & $\mathbf{ 28.45 }$& $23.62$ & $\mathbf{ 9.76 }$& $10.67$ \\ 
three\_agents & $9$ & $6$ & $500$ & $34.13$ & $\mathbf{ 1.70 }$& $1.81$ & $\mathbf{ 24.96 }$& $38.79$ & $35.81$ & $18.84$ & $\mathbf{ 6.35 }$& $6.47$ \\ 
triple\_waypoints & $20$ & $19$ & $1000$ & $10.76$ & $1.00$ & $\mathbf{ 0.56 }$& $\mathbf{ 36.44 }$& $170.02$ & $113.93$ & $26.04$ & $17.34$ & $\mathbf{ 15.73 }$\\ 
welding & $24$ & $25$ & $1000$ & $65.82$ & $1.93$ & $\mathbf{ 1.33 }$& $\mathbf{ 27.49 }$& $265.84$ & $185.70$ & $17.82$ & $14.49$ & $\mathbf{ 12.04 }$\\ 
handover & $12$ & $4$ & $100$ & $-$ & $\mathbf{ 0.80 }$& $2.85$ & $-$ & $52.30$ & $\mathbf{ 47.22 }$& $-$ & $\mathbf{ 12.12 }$& $12.13$ \\ 
eggcartons & $12$ & $19$ & $500$ & $5.12$ & $\mathbf{ 0.80 }$& $0.94$ & $\mathbf{ 60.05 }$& $164.81$ & $158.32$ & $55.51$ & $44.46$ & $\mathbf{ 43.27 }$\\ 
bottles & $12$ & $9$ & $500$ & $32.45$ & $\mathbf{ 16.00 }$& $92.72$ & $\mathbf{ 18.82 }$& $86.14$ & $59.69$ & $17.32$ & $15.16$ & $\mathbf{ 14.65 }$\\ 
box\_rearrangement & $12$ & $27$ & $1000$ & $14.10$ & $\mathbf{ 3.39 }$& $7.19$ & $\mathbf{ 38.30 }$& $223.80$ & $171.97$ & $32.98$ & $28.27$ & $\mathbf{ 27.11 }$\\ 
box\_reorientation & $12$ & $27$ & $500$ & $28.79$ & $\mathbf{ 3.76 }$& $4.95$ & $\mathbf{ 114.15 }$& $221.58$ & $207.76$ & $85.31$ & $55.74$ & $\mathbf{ 55.66 }$\\ 
pyramid & $24$ & $13$ & $500$ & $23.99$ & $3.23$ & $\mathbf{ 2.10 }$& $\mathbf{ 46.13 }$& $142.32$ & $105.08$ & $22.61$ & $18.09$ & $\mathbf{ 17.65 }$\\ 
box\_stacking & $24$ & $17$ & $1000$ & $14.98$ & $7.03$ & $\mathbf{ 4.51 }$& $\mathbf{ 53.34 }$& $204.13$ & $187.34$ & $26.99$ & $17.90$ & $\mathbf{ 16.68 }$\\ 
mobile\_wall\_four & $24$ & $17$ & $1000$ & $29.34$ & $\mathbf{ 11.61 }$& $13.38$ & $\mathbf{ 103.01 }$& $248.29$ & $220.33$ & $87.20$ & $46.73$ & $\mathbf{ 43.47 }$\\ 
mobile\_strut & $14$ & $91$ & $1000$ & $75.37$ & $\mathbf{ 13.67 }$& $49.35$ & $\mathbf{ 345.86 }$& $918.65$ & $747.70$ & $343.56$ & $326.51$ & $\mathbf{ 304.97 }$\\ 
three\_robot\_truss & $18$ & $29$ & $1000$ & $24.68$ & $3.92$ & $\mathbf{ 3.77 }$& $\mathbf{ 89.29 }$& $283.28$ & $211.93$ & $76.81$ & $69.99$ & $\mathbf{ 66.38 }$\\ 
spiral\_tower & $24$ & $31$ & $1000$ & $196.74$ & $\mathbf{ 26.37 }$& $31.00$ & $\mathbf{ 122.85 }$& $410.49$ & $311.52$ & $106.46$ & $100.44$ & $\mathbf{ 94.00 }$\\ 
spiral\_tower\_two & $12$ & $15$ & $500$ & $14.14$ & $\mathbf{ 1.37 }$& $2.02$ & $\mathbf{ 47.70 }$& $124.45$ & $110.74$ & $46.57$ & $44.21$ & $\mathbf{ 43.87 }$\\ 
cube\_four & $24$ & $49$ & $1000$ & $117.86$ & $\mathbf{ 41.87 }$& $59.44$ & $\mathbf{ 161.52 }$& $623.01$ & $423.71$ & $135.55$ & $135.30$ & $\mathbf{ 119.76 }$\\ 
\hline
dep\_piano & $6$ & $5$ & $500$ & $1.59$ & $\mathbf{ 0.41 }$& $0.61$ & $\mathbf{ 12.11 }$& $22.68$ & $20.92$ & $5.80$ & $\mathbf{ 3.55 }$& $3.55$ \\ 
dep\_box\_reorientation & $12$ & $27$ & $1000$ & $33.86$ & $\mathbf{ 5.92 }$& $7.29$ & $\mathbf{ 114.49 }$& $220.99$ & $206.70$ & $68.15$ & $57.14$ & $\mathbf{ 55.81 }$\\ 
dep\_box\_stacking & $24$ & $17$ & $1000$ & $20.31$ & $5.26$ & $\mathbf{ 2.93 }$& $\mathbf{ 60.03 }$& $205.04$ & $169.28$ & $26.49$ & $22.60$ & $\mathbf{ 20.05 }$\\ 
dep\_mobile\_wall\_four & $24$ & $17$ & $1000$ & $34.50$ & $22.70$ & $\mathbf{ 18.00 }$& $\mathbf{ 75.17 }$& $221.79$ & $182.59$ & $50.41$ & $44.18$ & $\mathbf{ 38.19 }$\\ 
\hline
unordered\_box\_reor. & $12$ & $14$ & $1000$ & $8.09$ & $16.43$ & $\mathbf{ 4.80 }$& $\mathbf{ 26.69 }$& $88.42$ & $83.23$ & $20.74$ & $21.48$ & $\mathbf{ 19.73 }$\\ 
unordered\_bottles & $12$ & $10$ & $500$ & $\mathbf{ 5.57 }$& $6.45$ & $7.07$ & $\mathbf{ 22.80 }$& $86.62$ & $67.38$ & $\mathbf{ 16.53 }$& $17.76$ & $17.15$ \\ 
\hline
unassigned\_tsp & $6$ & $6$ & $500$ & $0.97$ & $1.25$ & $\mathbf{ 0.10 }$& $\mathbf{ 6.32 }$& $14.90$ & $6.33$ & $2.60$ & $3.88$ & $\mathbf{ 2.49 }$\\ 
unassigned\_cleanup & $12$ & $12$ & $1000$ & $8.14$ & $30.36$ & $\mathbf{ 6.62 }$& $\mathbf{ 20.61 }$& $69.82$ & $66.76$ & $\mathbf{ 15.38 }$& $18.47$ & $17.24$ \\ 
unassigned\_stacking & $24$ & $10$ & $1000$ & $17.49$ & $110.49$ & $\mathbf{ 10.02 }$& $\mathbf{ 36.76 }$& $88.10$ & $93.85$ & $12.32$ & $11.84$ & $\mathbf{ 10.65 }$\\ 

\bottomrule
\end{tabular}
\end{table*}

\section{Discussion \& Limitations}
Due to planning in the composite space and treating the problem as a single big planning problem, the planners are not expected to scale well with the number of robots (due to the higher dimensionality) or with the number of goals, as can be seen in the slower convergence in the high dimensional problems.
However, we believe that formulating the problem in composite space helps understand how choices such as prioritization affect the continuous space, and this deeper understanding can enable the design of better planners.
\arxiv{We show more analysis on the scaling of the planners and where compute time is spent in \cref{sec:scaling,sec:compute}.}

We introduced a problem formulation and planners that can deal with all types of task specification.
However, we believe that there is a better way to approach the problem than a brute-force search on the task level, which the sampling-based planners effectively do here.
Despite these limitations, we find that the planning in composite space is likely good enough for many complex industrial settings, where the number of used robots does not go much beyond four robots.
Furthermore, with our initial mode sampling strategy defined in \cref{eq:prob_dist}, we were able to rapidly reduce the decision space by selecting a feasible mode sequence early in the search, allowing us to find an initial solution quickly and thus restricting the search space.

\section{Conclusion}
We presented a formalization of the multi-modal, multi-robot, multi-goal motion planning problem, and open-source the developed benchmark, which contains diverse problems reaching from simple environments with short goal sequences, to complex long horizon problems requiring intricate coordination of multiple robots and many mode switches.
We adapt multiple standard sampling-based planners that are probabilistically complete and asymptotically optimal to this type of problem, and show that the adaptions are crucial in order to obtain the shown performance. 
We also benchmarked a prioritized planner against these optimal planners and show that the solutions found by such a suboptimal planner can be competitive in some settings.

There are some clear improvements possible, such as exploiting the multi-robot structure of the planning problems, the nature of the mode families \cite{Kingston2021b}, or the multi-query-like structure of the problem \cite{hartmann2022eirm}.

There are many extensions to the chosen problem formulation: Both the problem formulation and the planners support extension of the transition logic to full multi-robot task and motion planning by changing the task-assignment-oracle.
In the future, we also want to support constrained multi-robot planning, or kinodynamic planning.
Finally, the formulation of the problem is amenable to parallelization, which could offer much faster planning times, and possibly better exploration of the problem space.

\section*{Acknowledgments}
{\small
The authors thank Miguel Zamora for insightful discussions and comments.
We would like to thank the reviewers for their comments.
This work was part of an Innovation project supported by Innosuisse.
VH was supported by the Swiss National Science Foundation through the National Centre of Competence in Digital Fabrication (Grant No. 200021 200644).
YH was supported by the ETH postdoc fellowship and the SNSF Ambizione Grant (Grant No. 223384).
}

\bibliographystyle{IEEEtran}
\bibliography{IEEEabrv,bib/bib}

\newpage
\appendix
\section*{Appendix}

\begin{figure*}[t]
    \begin{subfigure}[t]{.43\textwidth}
        \centering
        \includegraphics[width=.97\linewidth]{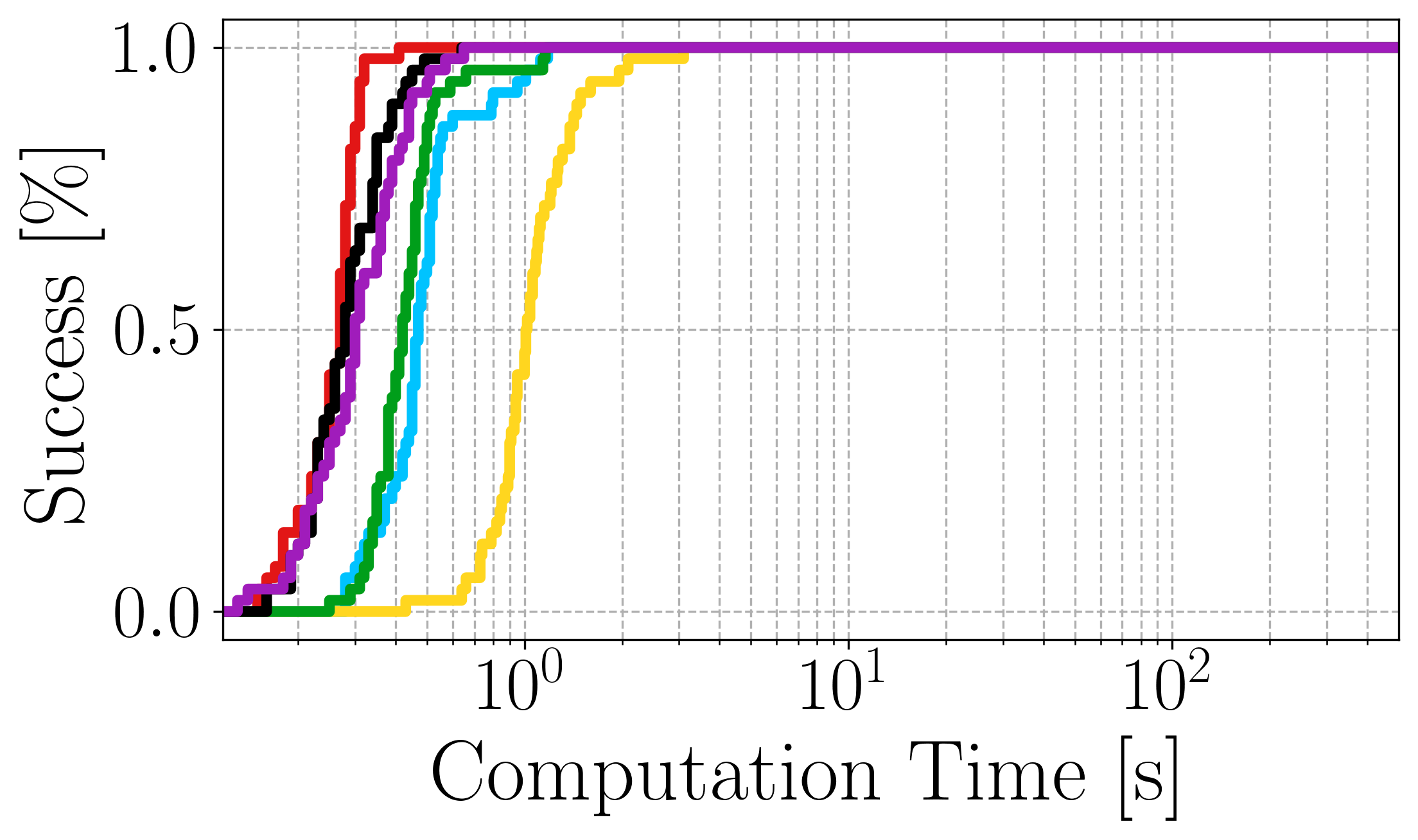}
        \caption{2D hallway (*).}
    \end{subfigure}\hfill
    \begin{subfigure}[t]{.43\textwidth}
        \centering
        \includegraphics[width=.97\linewidth]{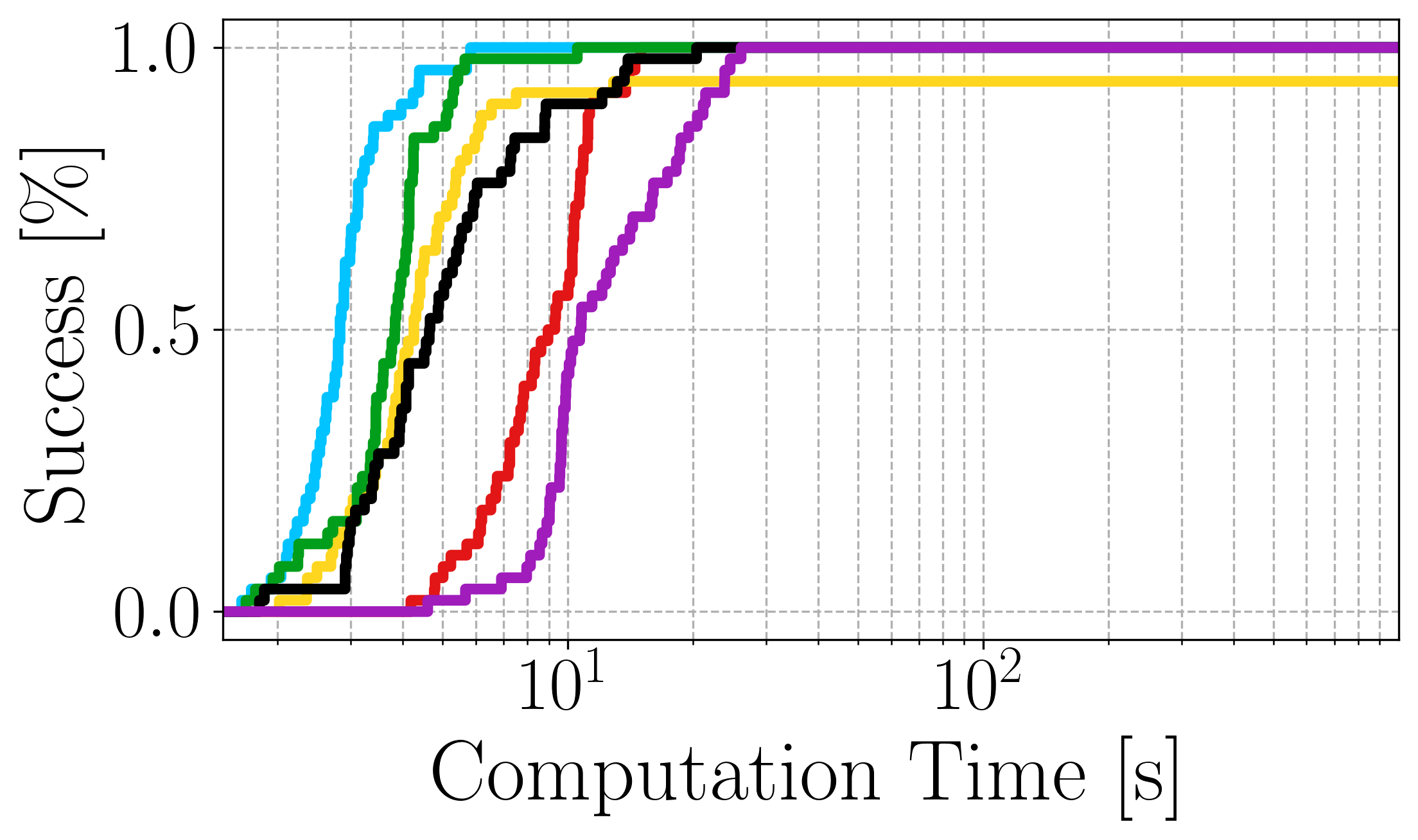}
        \caption{4-arm-box stacking (*).}
    \end{subfigure}\hfill
        \begin{subfigure}[t]{.43\textwidth}
        \centering
        \includegraphics[width=.97\linewidth]{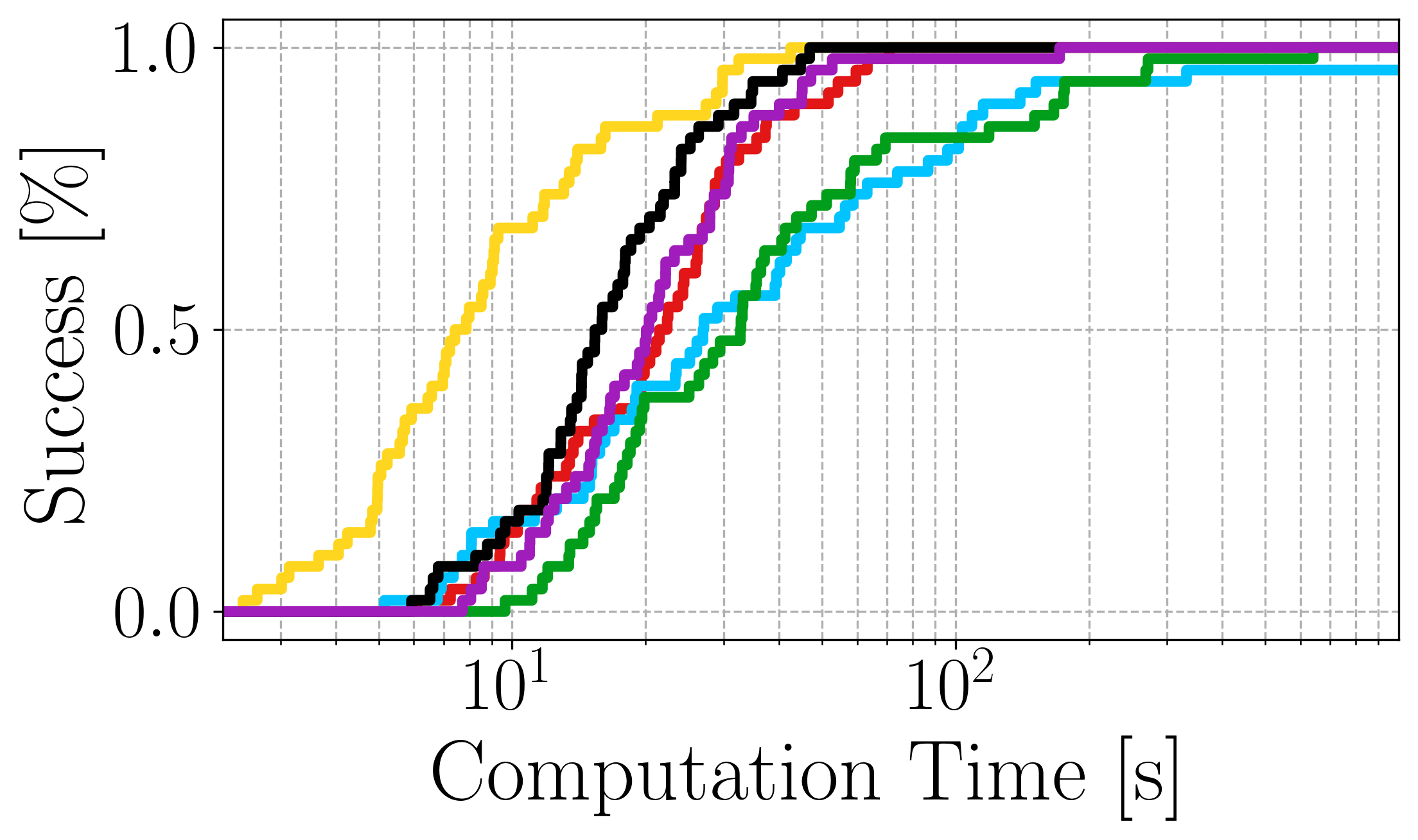}
        \caption{\label{fig:success_plots_mobile}Mobile assembly (**).}
    \end{subfigure}\hfill
    \begin{subfigure}[t]{.43\textwidth}
        \centering
        \includegraphics[width=.97\linewidth]{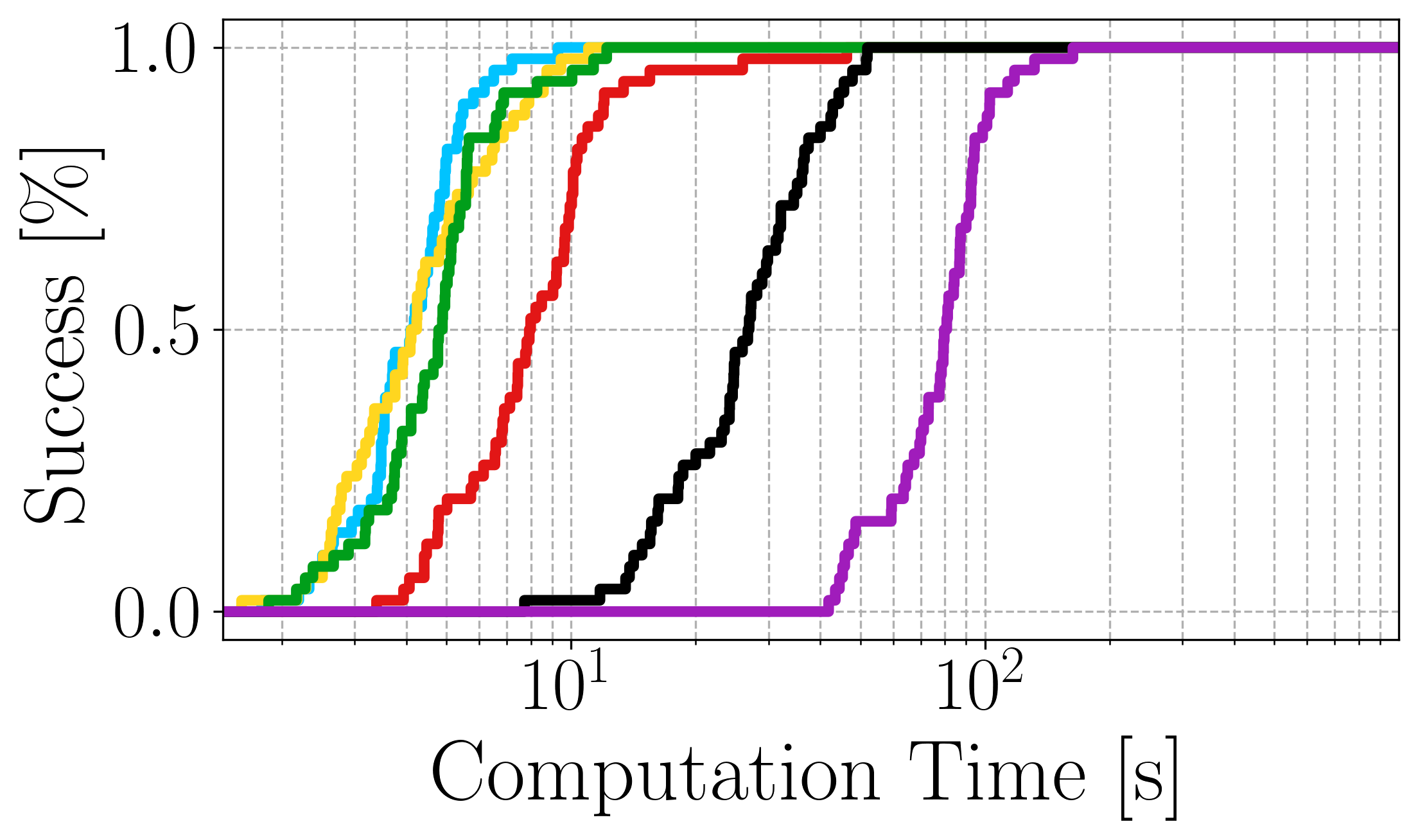}
        \caption{2-arm rearrangement (***).}
    \end{subfigure}
    \begin{subfigure}[b]{1.0\linewidth}%
        \centering
        \begin{tikzpicture}
\begin{axis} [
  width=\textwidth,
  height=0.5\textwidth,
  unbounded coords=jump,
  xtick align=inside,
  ytick align=inside,
  anchor=north,
  hide axis,
  xmajorgrids,
  ymajorgrids,
  major grid style={densely dotted, black!20},
  xmin=0,
  xmax=10,
  ymin=0,
  ymax=10,
  xlabel style={font=\footnotesize},
  xticklabel style={font=\footnotesize},
  ylabel style={font=\footnotesize},
  yticklabel style={font=\footnotesize},
  legend style={anchor=south, legend cell align=left, legend columns=6, at={(axis cs:5, 6)}, font=\small}
]
\addlegendimage{rrt, line width = 1.0pt, mark size=1.0pt, mark=square*}
\addlegendentry{RRT*}

\addlegendimage{birrt, line width = 1.0pt, mark size=1.0pt, mark=square*}
\addlegendentry{BiRRT*}

\addlegendimage{prm, line width = 1.0pt, mark size=1.0pt, mark=square*}
\addlegendentry{PRM*}

\addlegendimage{ait, line width = 1.0pt, mark size=1.0pt, mark=square*}
\addlegendentry{AIT*}

\addlegendimage{eit, line width = 1.0pt, mark size=1.0pt, mark=square*}
\addlegendentry{EIT*}

\addlegendimage{prioritized, line width = 1.0pt, mark size=1.0pt, mark=square*}
\addlegendentry{Prioritized}

\end{axis}
\end{tikzpicture}
    \end{subfigure}
    \caption{Evolution of the success rate time over 50 runs. (*) indicates a fully given task sequence, (**) indicates a partial task ordering, (***) indicates unassigned and unordered tasks. Note that the prioritized planner only acts as anytime planner and improves its cost in the (**) and (***) settings, where multiple different sequences can be generated and planned for.}
    \label{fig:success_plots}
\end{figure*}

\section{Success plots}
In addition to the cost convergence plots in \cref{fig:cost_plots}, we show the success rate of the planners over their runtime in \cref{fig:success_plots}.
The planners achieve a close to perfect success rate. 
The exception in this scenario here is the AIT* planner, which struggles with the mobile manipulation task (\cref{fig:success_plots_mobile}) which has a large configuration space and many tasks that are not fully ordered.
Thus, it can happen that the exploration of the large number of modes does not reach the goal mode, and does not find a solution.
In comparison, the other planners are more greedy, and thus do better in this scenario. 

\section{Scaling studies}
\label{sec:scaling}
\begin{figure*}[t]
    \begin{subfigure}[t]{.43\textwidth}
        \centering
        \includegraphics[width=.97\linewidth]{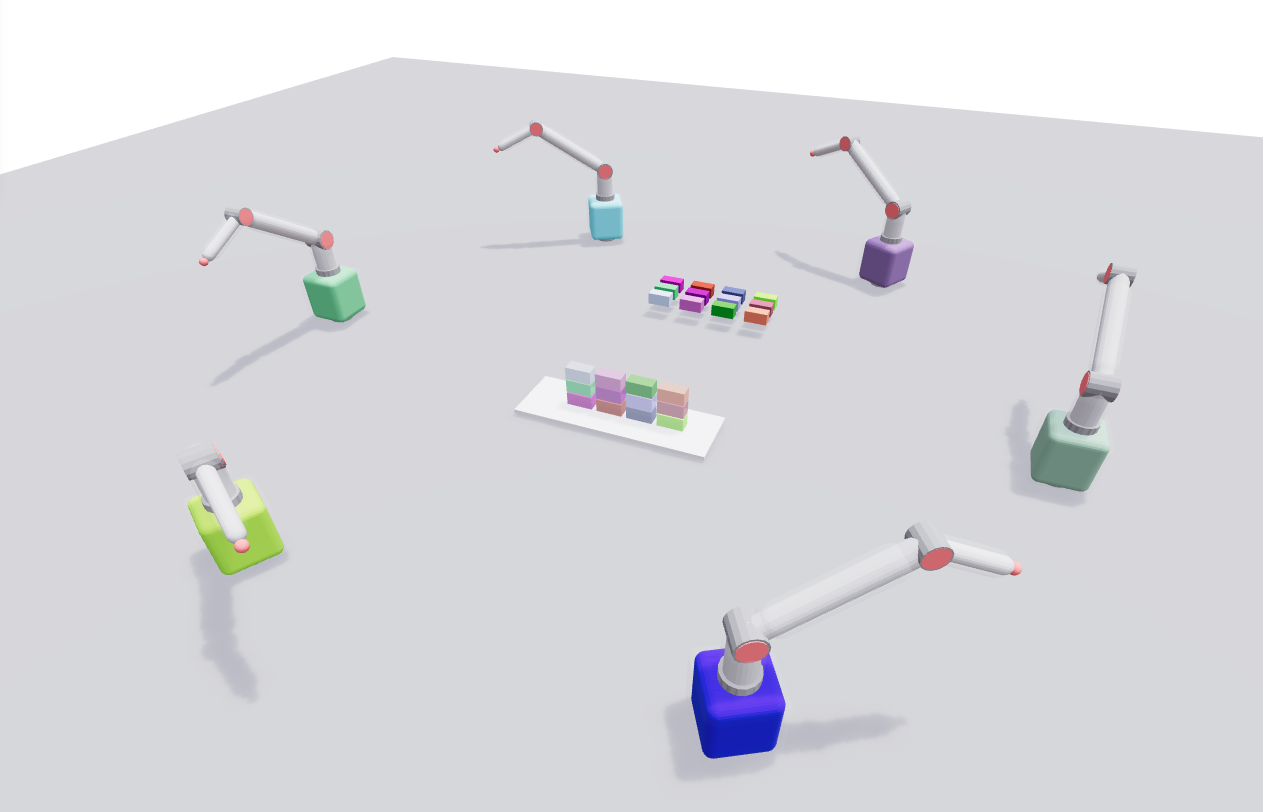}
    \end{subfigure}\hfill
    \begin{subfigure}[t]{.43\textwidth}
        \centering
        \includegraphics[width=.97\linewidth]{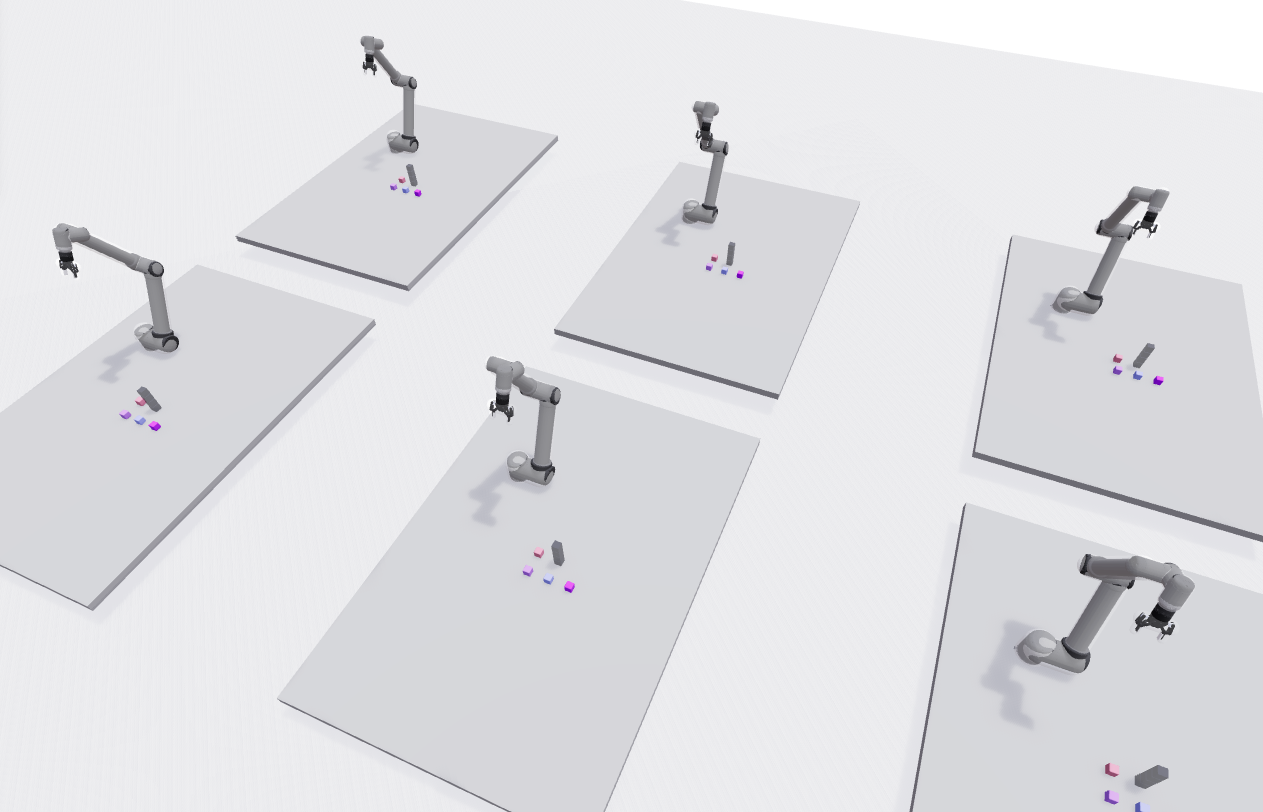}
    \end{subfigure}
    \caption{Scenarios for the empirical scaling analysis. Left: Mobile manipulation with 6 robots, right: independent box stacking with 6 robots.}
    \label{fig:scaling_scenarios}
\end{figure*}

We present an empirical analysis of how the proposed algorithms scale with the number of robots and the number of tasks on the example of the bidirectional RRT.
We show the scaling behavior of the planner for three scenarios:
\begin{itemize}
    \item Box stacking with multiple robots in the same workspace, i.e., the scenario shown in \cref{fig:intro_image}.
    \item Multiple robots building a wall, with up to 7 robots, i.e., with up to 42 Dof, and up to 25 bricks (\cref{fig:scaling_scenarios}, left).
    \item Box stacking of multiple completely independent robots in independent workspaces, i.e., the box stacking scenario in \cref{fig:intro_image} with a single robot, duplicated $n$ times (\cref{fig:scaling_scenarios}, right).
\end{itemize}
In these scenarios, we focus on the time the planner takes to reach the initial solution dependent on the number of robots in the scene, and dependent on the number of tasks.
We do not analyze the convergence behavior of the planner here.

\subsection{Scaling in shared workspaces}
\begin{figure*}[t]
    \begin{subfigure}[t]{.43\textwidth}
        \centering
        \includegraphics[width=.97\linewidth]{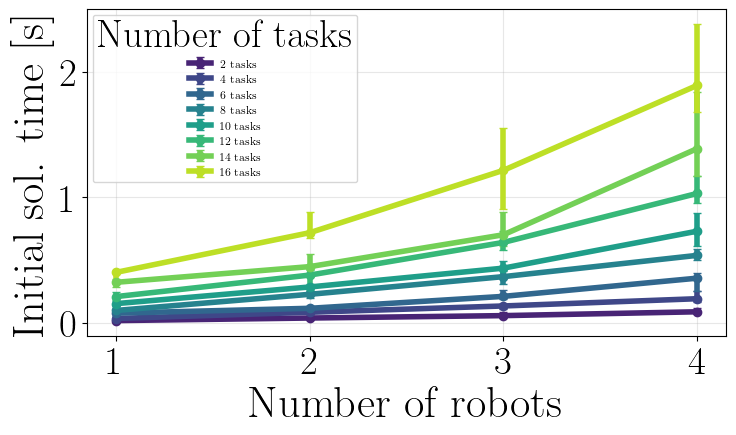}
        \caption{\label{fig:box_scaling_robots}Box stacking: Scaling with \#robots.}
    \end{subfigure}\hfill
    \begin{subfigure}[t]{.43\textwidth}
        \centering
        \includegraphics[width=.97\linewidth]{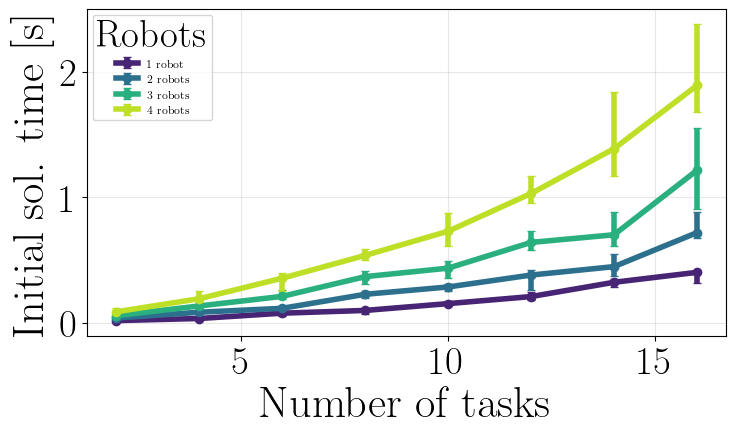}
        \caption{\label{fig:box_scaling_tasks}Box stacking: Scaling with \#tasks.}
    \end{subfigure}\hfill
        \begin{subfigure}[t]{.43\textwidth}
        \centering
        \includegraphics[width=.97\linewidth]{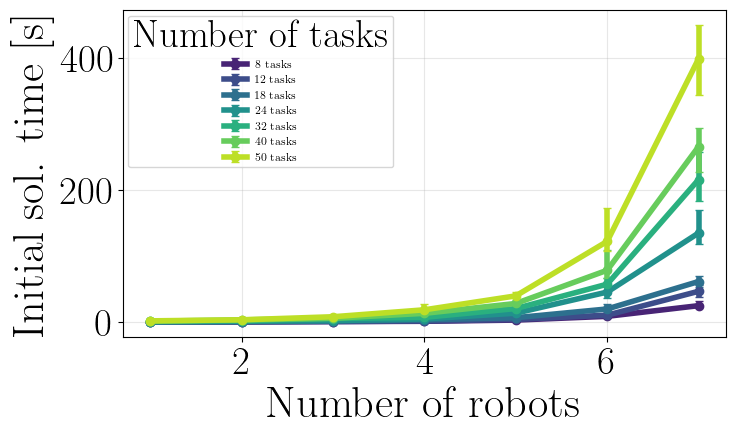}
        \caption{\label{fig:mobile_scaling_robots}Mobile wall: Scaling with \#robots.}
    \end{subfigure}\hfill
    \begin{subfigure}[t]{.43\textwidth}
        \centering
        \includegraphics[width=.97\linewidth]{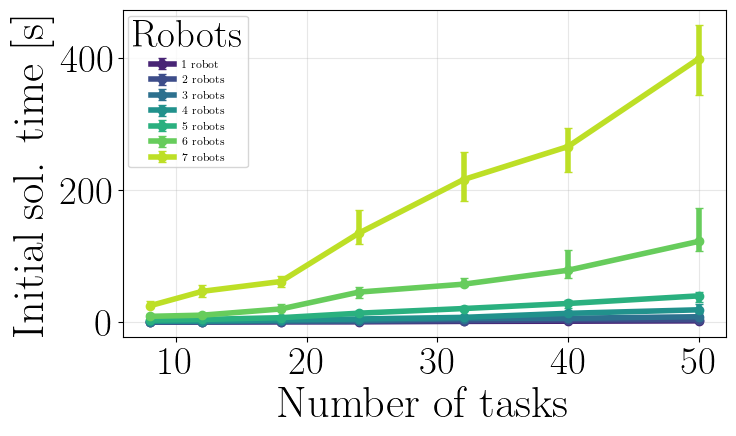}
        \caption{\label{fig:mobile_scaling_tasks}Mobile wall: Scaling with \#tasks.}
    \end{subfigure}
    \caption{Median time taken by the bidirectional RRT to find the initial solution in the box stacking and the mobile manipulation setting in shared workspaces along with the confidence intervals.}
    \label{fig:scaling_shared}
\end{figure*}
In order to analyze the scaling behavior, we plot the number of robots against the time taken to find the initial solution (for varying numbers of tasks) in \cref{fig:box_scaling_robots,fig:mobile_scaling_robots}, and we plot the number of tasks against the time taken to find an initial solution (for varying numbers of robots) in \cref{fig:box_scaling_tasks,fig:mobile_scaling_tasks}.

We can see that similar trends hold for both the environment with the tabletop box stacking and for the environment with the mobile robots: in both scenarios, we see slightly above linear complexity with the number of tasks, and we see roughly exponential complexity with the number of robots.
This is more clearly visible in the scenario with the mobile robots where we have both more robots, and more tasks to note the differences.

\subsection{Scaling in independent workspaces}
\begin{figure*}[t]
    \begin{subfigure}[t]{.43\textwidth}
        \centering
        \includegraphics[width=.97\linewidth]{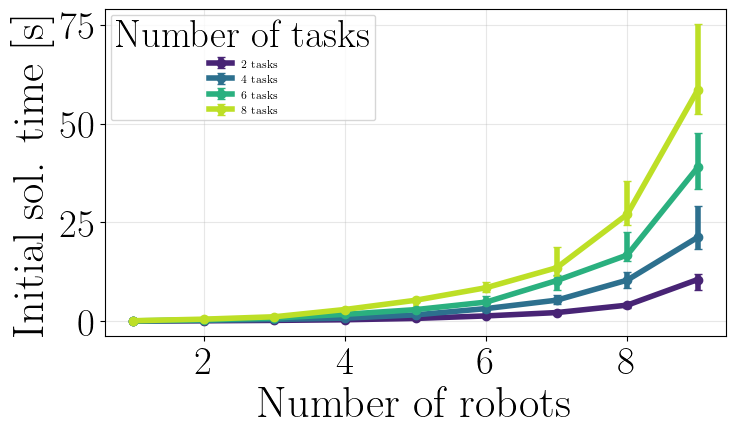}
        \caption{\label{fig:independent_box_scaling_robots}Independent box stacking: Scaling with \#robots.}
    \end{subfigure}\hfill
    \begin{subfigure}[t]{.43\textwidth}
        \centering
        \includegraphics[width=.97\linewidth]{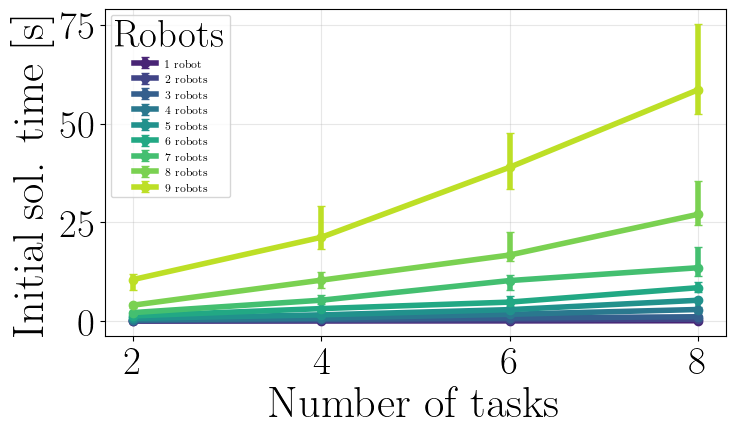}
        \caption{\label{fig:independent_box_scaling_tasks}Independent box stacking: Scaling with \#tasks.}
    \end{subfigure}
    \caption{Median time taken by the bidirectional RRT to find the initial solution in the box stacking setting in independent workspaces along with the confidence intervals.
    Note that in this scenario, we are considering \textit{tasks per robot} instead of the absolute number of tasks as before.}
    \label{fig:scaling_independent}
\end{figure*}
For the last scenario where each robot is completely independent of the others, we group by number of tasks \textit{per robot}, as it is not easily possible to hold the number of tasks constant in the isolated environments when varying the number of robots.
In this problem, an ideal planner would not be influenced by the presence of the other environments (except through the ordering constraint of the tasks), and the time taken for planning would be expected to increase linearly with the number of environments (or stay constant if we could fully parallelize everything).
In \cref{fig:scaling_independent}, we can see that this is not the case for the composite space planning approach.
Rather, we see similar behavior as in the previous environments, i.e., the roughly exponential increase of time taken to find the initial solution.

\section{Compute breakdown}
\label{sec:compute}

We provide a few examples of where compute time is spent in the bidirectional RRT.
Particularly, we show the fractions of where the compute time is spent until an initial solution is found for all the environments also used in \cref{tab:all_exp}. 
We show the times here up to where the initial solution is found only, as afterwards, the profile of where the time is spent changes considerably, and is more dependent on the hyperparameters, e.g., deciding on how much shortcutting we do.

\begin{table}[t]
\scriptsize
    \centering
\caption{Ratios of where time is spent in the bidirectional RRT planner up until the initial solution is found.
\textit{Rnd. Sampling} contains collision checking time as well.
Major parts in \textit{Other} are nearest neighbor computation, oracle evaluation and partially overhead from, e.g., bookkeeping for modes.}
    \label{tab:fraction_compute_use}
\begin{tabular}{l | r@{\hspace{8pt}} r@{\hspace{8pt}} r@{\hspace{8pt}} r@{\hspace{8pt}} r}
\toprule
& \multicolumn{5}{c}{Fraction of total time [\%]} \\
\cmidrule(lr){2-6}
& & \multicolumn{3}{c}{Collision checking} & \\
\cmidrule(lr){3-5}
Environment& Rnd. Sampling & Single config & Edge (free) & Edge (blocked) & Other \\
\midrule
simple & $15.3$ & $2.0$ & $\mathbf{54.1}$ & $11.2$ & $17.4$ \\
one\_agent\_many\_goals & $16.6$ & $0.5$ & $\mathbf{55.3}$ & $0.8$ & $26.9$ \\
single\_agent\_mover & $19.1$ & $2.0$ & $\mathbf{44.4}$ & $7.4$ & $27.2$ \\
2d\_handover & $23.7$ & $5.2$ & $\mathbf{49.4}$ & $10.4$ & $11.2$ \\
random\_2d & $\mathbf{51.6}$ & $4.3$ & $17.1$ & $12.1$ & $14.9$ \\
other\_hallway & $22.8$ & $6.3$ & $\mathbf{39.5}$ & $7.6$ & $23.8$ \\
three\_agents & $\mathbf{36.2}$ & $3.2$ & $26.1$ & $14.3$ & $20.2$ \\
triple\_waypoints & $7.6$ & $0.6$ & $\mathbf{83.8}$ & $2.4$ & $5.6$ \\
welding & $11.2$ & $0.6$ & $\mathbf{82.1}$ & $1.6$ & $4.7$ \\
handover & $\mathbf{37.1}$ & $7.5$ & $25.1$ & $13.0$ & $17.2$ \\
eggcartons & $10.3$ & $1.1$ & $\mathbf{70.7}$ & $3.2$ & $14.8$ \\
bottles & $1.3$ & $0.3$ & $\mathbf{95.3}$ & $2.6$ & $0.5$ \\
box\_rearrangement & $17.6$ & $1.5$ & $\mathbf{64.3}$ & $4.1$ & $12.4$ \\
box\_reorientation & $20.4$ & $2.1$ & $\mathbf{60.9}$ & $5.2$ & $11.5$ \\
pyramid & $20.7$ & $0.7$ & $\mathbf{70.6}$ & $2.5$ & $5.5$ \\
box\_stacking & $31.7$ & $1.1$ & $\mathbf{52.6}$ & $2.7$ & $12.0$ \\
mobile\_wall\_four & $18.1$ & $8.0$ & $\mathbf{48.2}$ & $17.2$ & $8.5$ \\
mobile\_strut & $3.7$ & $0.9$ & $\mathbf{85.3}$ & $2.8$ & $7.3$ \\
spiral\_tower & $28.3$ & $1.5$ & $\mathbf{61.2}$ & $3.5$ & $5.6$ \\
spiral\_tower\_two & $6.8$ & $0.9$ & $\mathbf{80.5}$ & $3.4$ & $8.4$ \\
cube\_four & $35.8$ & $1.4$ & $\mathbf{56.9}$ & $3.1$ & $2.8$ \\
\hline
dep\_piano & $29.6$ & $2.7$ & $\mathbf{39.2}$ & $9.8$ & $18.7$ \\
dep\_box\_stacking & $31.7$ & $0.9$ & $\mathbf{55.5}$ & $2.1$ & $9.8$ \\
dep\_box\_reorientation & $14.6$ & $1.3$ & $\mathbf{74.8}$ & $3.1$ & $6.2$ \\
dep\_mobile\_wall\_four & $22.1$ & $7.9$ & $\mathbf{38.1}$ & $17.5$ & $14.4$ \\
\hline
unordered\_box\_reorientation & $11.9$ & $0.8$ & $\mathbf{64.9}$ & $1.6$ & $20.8$ \\
unordered\_bottles & $8.1$ & $1.9$ & $\mathbf{76.3}$ & $4.5$ & $9.2$ \\
\hline
unassigned\_tsp & $17.5$ & $0.3$ & $\mathbf{50.9}$ & $0.9$ & $30.4$ \\
unassigned\_cleanup & $11.3$ & $0.6$ & $\mathbf{59.3}$ & $0.9$ & $27.9$ \\
unassigned\_stacking & $18.9$ & $0.5$ & $\mathbf{60.5}$ & $1.1$ & $18.9$ \\
\bottomrule
\end{tabular}

\end{table}

We can see in \cref{tab:fraction_compute_use} that up to the initial solution, the dominating factor is edge collision checking in most problems.
We want to note that it is in particular the valid edges that take the majority of the time, and not the invalid edges.

The ratio of the main time-consuming functions is also dependent on the environment, and the dimensionality of the problem:
The probability of sampling a collision free configuration decreases with increasing dimensionality of the problem space.
Thus, the ratio of edge-collision checking to configuration sampling also changes depending on this.

After the planner found the initial solution, i.e., when improving the path using informed sampling and shortcutting, more time is spent doing edge-collision checking (from rewiring and from shortcutting) and configuration-collision checking (from informed sampling).
The ratio of what dominates here is very dependent on the hyperparameters, since we decide how many iterations the planner does for shortcutting, and how long it tries to generate informed samples.

\subsection{Scaling of compute ratios}
We further break down where time is spent if we increase the number of robots in the independent box stacking environment, and in the mobile manipulation scenario from the previous section.
We focus on the time taken to find the initial solution, and not on the convergence.
In order to simplify the analysis, we focus on only one specific set of tasks, i.e. we keep the number of tasks per robot constant over all the experiments.

\begin{figure*}[t]
    \begin{subfigure}[t]{.43\textwidth}
        \centering
        \includegraphics[width=.97\linewidth]{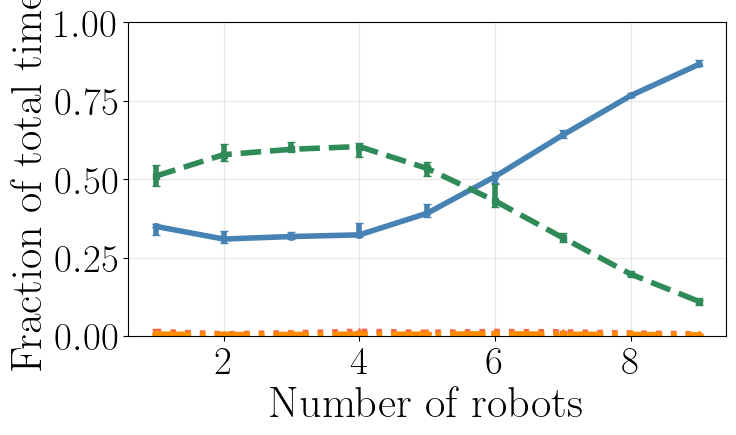}
    \end{subfigure}\hfill
    \begin{subfigure}[t]{.43\textwidth}
        \centering
        \includegraphics[width=.97\linewidth]{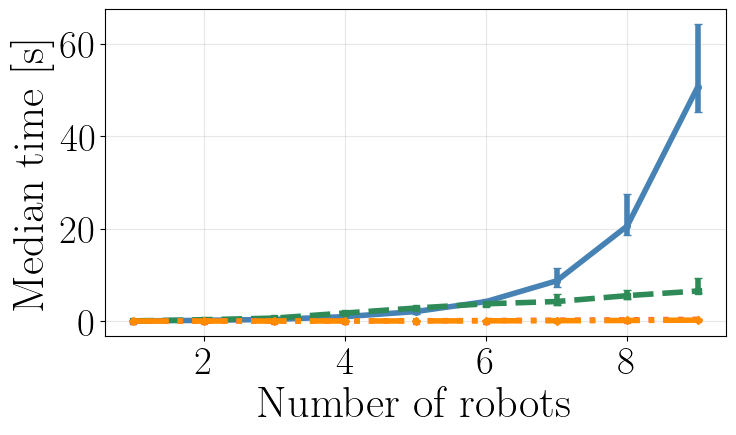}
    \end{subfigure}
    \begin{subfigure}[b]{1.0\linewidth}%
        \centering
        \begin{tikzpicture}
\begin{axis} [
  width=\textwidth,
  height=0.5\textwidth,
  unbounded coords=jump,
  xtick align=inside,
  ytick align=inside,
  anchor=north,
  hide axis,
  xmajorgrids,
  ymajorgrids,
  major grid style={densely dotted, black!20},
  xmin=0,
  xmax=10,
  ymin=0,
  ymax=10,
  xlabel style={font=\footnotesize},
  xticklabel style={font=\footnotesize},
  ylabel style={font=\footnotesize},
  yticklabel style={font=\footnotesize},
  legend style={anchor=south, legend cell align=left, legend columns=6, at={(axis cs:5, 6)}, font=\small}
]
\addlegendimage{rnd_sample, line width = 1.0pt, mark size=1.0pt, mark=square*}
\addlegendentry{Rnd. Sampling}

\addlegendimage{edge_valid, line width = 1.0pt, mark size=1.0pt, mark=square*}
\addlegendentry{Edge check (valid)}

\addlegendimage{edge_invalid, line width = 1.0pt, mark size=1.0pt, mark=square*}
\addlegendentry{Edge check (invalid)}

\addlegendimage{conf_check, line width = 1.0pt, mark size=1.0pt, mark=square*}
\addlegendentry{Conf. check}

\end{axis}
\end{tikzpicture}
    \end{subfigure}
    \caption{Independent box stacking: Relative (left) and absolute (right) compute time spent by the bidirectional RRT in different functions depending on the number of robots in the environment.}
    \label{fig:indep_scaling_compute}
\end{figure*}

\begin{figure*}[t]
    \begin{subfigure}[t]{.43\textwidth}
        \centering
        \includegraphics[width=.97\linewidth]{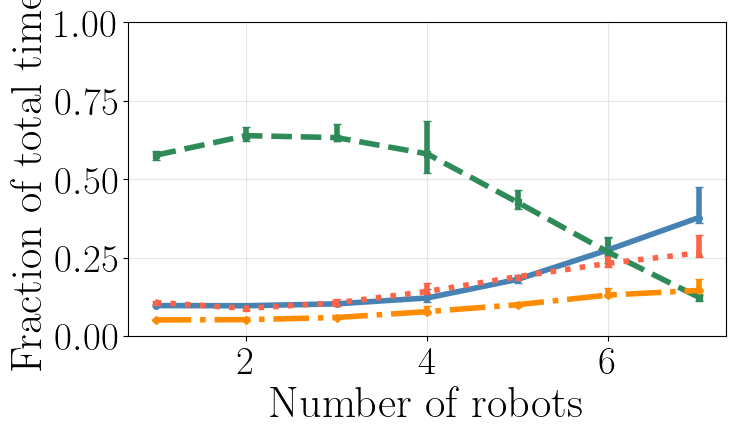}
    \end{subfigure}\hfill
    \begin{subfigure}[t]{.43\textwidth}
        \centering
        \includegraphics[width=.97\linewidth]{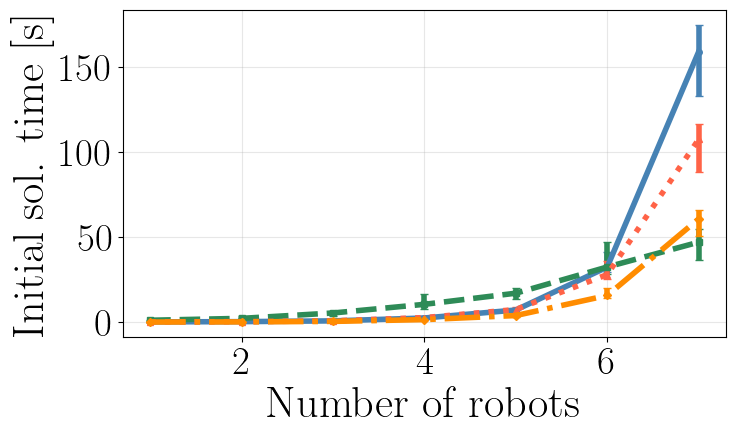}
    \end{subfigure}
    \begin{subfigure}[b]{1.0\linewidth}%
        \centering
        \begin{tikzpicture}
\begin{axis} [
  width=\textwidth,
  height=0.5\textwidth,
  unbounded coords=jump,
  xtick align=inside,
  ytick align=inside,
  anchor=north,
  hide axis,
  xmajorgrids,
  ymajorgrids,
  major grid style={densely dotted, black!20},
  xmin=0,
  xmax=10,
  ymin=0,
  ymax=10,
  xlabel style={font=\footnotesize},
  xticklabel style={font=\footnotesize},
  ylabel style={font=\footnotesize},
  yticklabel style={font=\footnotesize},
  legend style={anchor=south, legend cell align=left, legend columns=6, at={(axis cs:5, 6)}, font=\small}
]
\addlegendimage{rnd_sample, line width = 1.0pt, mark size=1.0pt, mark=square*}
\addlegendentry{Rnd. Sampling}

\addlegendimage{edge_valid, line width = 1.0pt, mark size=1.0pt, mark=square*}
\addlegendentry{Edge check (valid)}

\addlegendimage{edge_invalid, line width = 1.0pt, mark size=1.0pt, mark=square*}
\addlegendentry{Edge check (invalid)}

\addlegendimage{conf_check, line width = 1.0pt, mark size=1.0pt, mark=square*}
\addlegendentry{Conf. check}

\end{axis}
\end{tikzpicture}
    \end{subfigure}
    \caption{Mobile manipulation: Relative (left) and absolute (right) compute time spent by the bidirectional RRT in different functions depending on the number of robots in the environment.}
    \label{fig:mobile_scaling_compute}
\end{figure*}

In the following, we distinguish (i) time spent in sampling random valid configurations (which includes collision checking them, (ii) collision checking single configurations (as done, e.g., after steering), and (iii) edge collision checking, which we further break down into valid and invalid ones based on the checked results.

In the independent box stacking scenario, the environment per robot does not change with the addition of more robots.
Thus, we would expect the sampling, and the collision checking effort to increase roughly linearly with the number of robots.
However, as we can see in \cref{fig:indep_scaling_compute,fig:mobile_scaling_compute}, the time spent in sampling of valid configurations, and edge-collision checking is increasing considerably with the number of robots in the BiRRT* planner.

Focusing on the independent box stacking environment, we can observe two main effects:
\begin{itemize}
    \item The time spent finding valid random configurations in this particular environment scales exponentially.
    \item Edge collision checking effort increases significantly. We attribute this to the fact that the expected path length for an edge increases with the number of robots. 
\end{itemize}

In the mobile manipulation scenario, we can see that the time spent in checking edges that are invalid increases with the number of robots, which is likely mostly due to the more congested environment due to more robots being in the same workspace. 

Summarizing everything, we can see that the time to find an initial solution increases roughly exponentially with the number of robots that we plan for, which can be expected, as the volume of the search space grows exponentially with the number of degrees of freedom.

\textbf{Sampling collision free poses:} We notice that one of the large drivers of the scaling is sampling collision free poses.
In hindsight, this can be expected, given that the probability of sampling a collision-free pose in the full composite space is roughly $p_{\text{free}}\propto \prod_i p_{\text{free},i}$, where we use $p_{\text{free},i}$ as probability of sampling a collision free pose for robot $i$.

\textbf{Edge collision checking:} While we would expect the robots to roughly maintain the same path length between their subsequent goals independent of how many other robots are added (especially in the independent environments), we can observe that the paths that the robots take between their goals become longer the more other robots we add to the problem.
This leads to more collision checking to validate the motions of the robots, and therefore a longer time until we find the initial solution to the problem.

\subsection{Tackling the scaling effects}
The described issues can be alleviated by (partially) decoupling the planning of the different robots from each other, as done in many previous works.
However, completely decoupling the robots from each other sacrifices completeness and optimality.
Luckily, one does not have to decouple everything completely: It is for example possible to sample full configurations more efficiently via Gibbs sampling, or simply by sampling them iteratively instead of all at once.
This brings us from the exponential, to a roughly linear complexity, which we would also have if we plan in a completely decoupled space, where we treat previously planned robots as fixed.

The second issue (the growing length of the paths between goals) can also be alleviated, e.g., by biasing the paths that we plan towards good single-robot plans, which again goes in the direction of decoupling the robots from each other.

\end{document}